\documentclass[11pt]{article}
\usepackage[final]{acl}
\usepackage{times}
\usepackage{latexsym}

\definecolor{cGrey}{HTML}{efefef} %
\definecolor{cGreen}{HTML}{0077b6} %
\usepackage{xcolor}
\definecolor{mjgreen}{rgb}{0.0,0.5,0.0}
\definecolor{teal}{rgb}{0.0,0.5,0.5}

\definecolor{Blue}{HTML}{ecf4f8} %

\usepackage{colortbl}
\usepackage{bm}
\usepackage{multirow}
\usepackage{amsmath}
\usepackage{pifont}
\usepackage{booktabs}
\usepackage{amssymb}
\usepackage{tipa}
\usepackage{booktabs, multirow, xcolor, colortbl, array}
\usepackage{makecell}
\usepackage{colortbl}
\usepackage{tipa}
\usepackage{xspace}
\usepackage[T5,T3, T1]{fontenc}    %
\usepackage{CJKutf8}

\usepackage{tabularray}
\UseTblrLibrary{booktabs}
\usepackage{enumitem}

\AtEndPreamble{
    \usepackage[capitalize]{cleveref}
    \crefname{section}{Sec.}{Secs.}
    \Crefname{section}{Section}{Sections}
    \Crefname{table}{Table}{Tables}
    \crefname{table}{Tab.}{Tabs.}
    \Crefname{figure}{Figure}{Figures}
}

\definecolor{mjgreen}{HTML}{228B22}

\usepackage[utf8]{inputenc}
\usepackage{microtype}
\usepackage{inconsolata}
\usepackage{graphicx}

\newcommand{\vn}[1]{{\fontencoding{T5}\selectfont #1}}
\newcommand{\ko}[1]{\begin{CJK}{UTF8}{mj}#1\end{CJK}}
\newcommand{\benchmark}{\textsc{UGTPhon}\xspace}
\newcommand{\Gnc}{\mathrm{G}_{nc}}
\newcommand{\Gc}{\mathrm{G}_c}
\newcommand{\Pc}{\mathrm{P}_c}

\title{Phonemizing User-Generated Text: \\ A Benchmark, Taxonomy, and Compositional Approach}
\newcommand{\PERc}{\text{PER}_{c}}
\newcommand{\PERnc}{\text{PER}_{nc}}
\newcommand{\PERo}{\text{PER}_{all}}
\newcommand{\dn}{$\scriptscriptstyle\downarrow$}

\author{
  \textbf{MinJu Jeon\textsuperscript{1,2}}\thanks{
    Work performed during an internship at NAVER Cloud.
  },
  \textbf{Younghan Park\textsuperscript{3}},
  \textbf{Han Sung Park\textsuperscript{4}},
  \textbf{Jong-Hwan Kim\textsuperscript{1}},
\\
  \textbf{Dong-Jin Kim\textsuperscript{2}}\thanks{
    Corresponding authors.
  },
  \textbf{Hoyeon Lee\textsuperscript{1}}\footnotemark[2]
\\
\\
  \textsuperscript{1}NAVER Cloud,
  \textsuperscript{2}Hanyang University,
\\
  \textsuperscript{3}Carnegie Mellon University,
  \textsuperscript{4}Georgia Institute of Technology
}

\begin{document}
\maketitle
\begin{abstract}
Text-to-speech systems increasingly process user-generated text (UGT) such as \textit{ppl} and \textit{imo}, whose pronunciation must be inferred from the canonical rather than surface form.
We introduce \benchmark, the first grapheme-to-phoneme (G2P) benchmark for UGT in English, Vietnamese, and Korean, together with an inference-grounded taxonomy for fine-grained diagnosis.
Existing G2P models and frontier LLMs exhibit a systematic canonical-to-non-canonical performance gap, reaching up to $66.8$ PER points.
As a benchmark baseline, we propose a simple compositional G2P approach that incorporates canonical-form evidence through exact-match lookup and staged decoding. Across matched ByT5 and Qwen2.5-0.5B backbones, explicit canonical-form modeling consistently reduces non-canonical G2P errors. The 0.5B variant also performs competitively with much larger few-shot frontier LLMs, highlighting the benefit of explicitly modeling canonical-form inference for UGT phonemization.
\end{abstract}

\section{Introduction}
Grapheme-to-phoneme (G2P) conversion~\cite{bisani2008joint, novak2012wfst, rao2015grapheme} maps a word's written form to its phonemic transcription. It serves as the essential front-end of text-to-speech (TTS) systems~\cite{ren2019fastspeech, ploujnikov2022soundchoice}, where synthesis quality depends directly on the pronunciations it produces~\cite{taylor2005hidden,van2017expanded}. Existing G2P approaches, including joint-sequence models~\cite{bisani2008joint}, transformer architectures~\cite{yolchuyeva2020transformer}, and pretrained byte-level models~\cite{byt5g2p}, achieve robust performance on canonical lexicons such as IPA-Dict~\cite{ipadict} and WikiPron~\cite{wikipron}. TTS pipelines, therefore, treat phonemization as a reliable upstream module~\cite{kim2021conditional, shen2018natural}, assuming input that follows standard orthography.

\begin{figure}[t]
\begin{center}
\includegraphics[width=\linewidth]{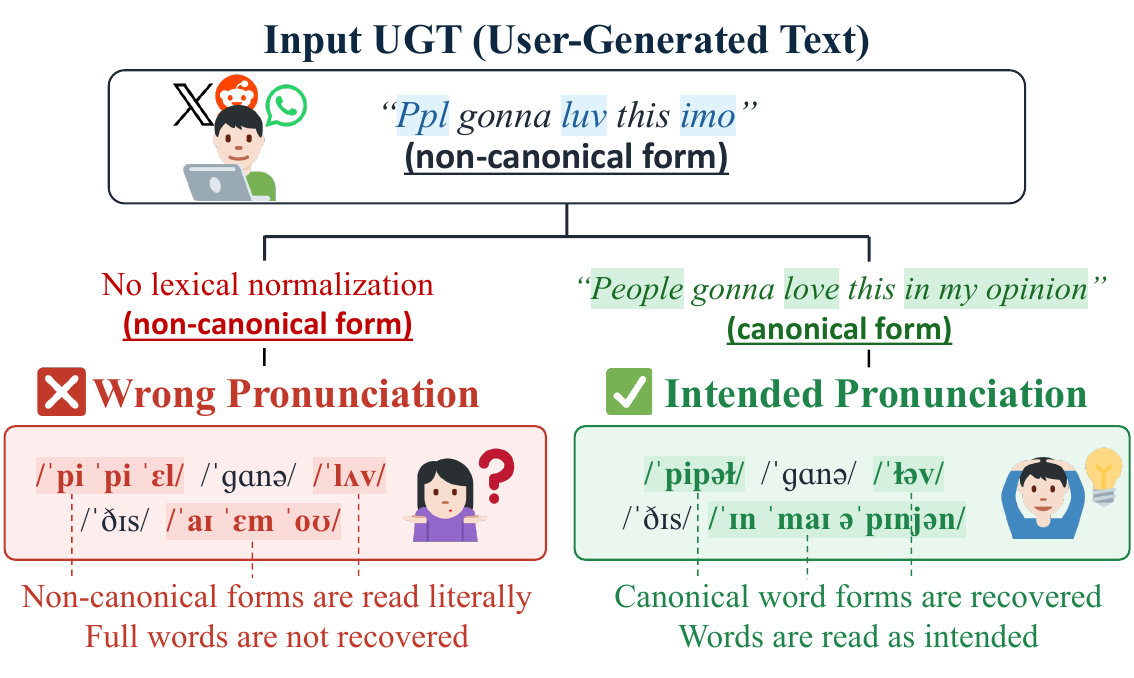}
\end{center}
\caption{Conventional G2P models (red, left) read non-canonical words (\textit{ppl}, \textit{luv}, \textit{imo}) letter by letter, producing letter names. Accurate phonemization (green, right) therefore requires recovering the canonical form (\textit{people}, \textit{love}, \textit{in my opinion}) before generating phonemes.}
\label{fig:problem}
\end{figure}

However, this assumption breaks down in practice. As TTS systems increasingly process conversational and web-based text~\cite{lee2022impact}, input is dominated by \emph{user-generated text} (UGT)~\cite{plank2016non, han2011lexical}: a mixture of canonical words (e.g., \textit{love}, \textit{people}) and non-canonical forms like \textit{luv} and \textit{ppl}. These non-canonical forms follow systematic patterns recurring across languages such as English (EN), Vietnamese (VI), and Korean (KO): abbreviations (EN:~\textit{lmao}, VI:~\vn{\textit{vcl}}, KO:~\ko{ㅇㄱㄹㅇ}), orthographic lengthening (\textit{cooool}, \vn{\textit{điiiii}}, \ko{ㅋㅋㅋㅋㅋㅋ}), phonetic substitutions (\textit{gr8}, \vn{\textit{wá}}, \ko{머해}), and slang (\textit{lit}, \vn{\textit{gato}}, \ko{킹받다}). Existing G2P models therefore read these forms literally and yield letter-name strings rather than the intended pronunciations (\Cref{fig:problem} (A)). Prior work has framed UGT as noise~\citep{zhao2022r}, but UGT is fundamentally different: \textit{imo} is a deliberate substitution whose pronunciation must be inferred from its canonical expansion (\textit{in my opinion}), not its surface form. G2P robustness under UGT remains unexplored.

To address this gap, we introduce \benchmark\footnote{The \benchmark is publicly available at \url{https://github.com/naver-ai/UGTPHON}.}, the first G2P benchmark for UGT, spanning three typologically diverse languages (English, Vietnamese, and Korean) with the intended canonical form and gold phoneme sequence for every non-canonical word.

To support fine-grained diagnosis, we further group non-canonical words by the inference a G2P model must perform: \textbf{(1) Reconstruct} (recover dropped characters, e.g., \textit{idk} for \textit{I don't know}), \textbf{(2) Repetition} (collapse prosodic repetitions, e.g., \textit{soooo} for \textit{so}), and \textbf{(3) Pass-through} (the surface already encodes the pronunciation, e.g., \textit{luv} for \textit{love}). This inference-grounded design lets a model's failure pattern indicate which kind of inference is missing, not which inputs it fails on.

Non-canonical G2P composes two stages, inferring the canonical form and phonemizing it. Standard end-to-end training collapses these into one, while a naive normalize-then-G2P pipeline over-corrects forms that need no rewriting. We therefore use a \textbf{simple compositional G2P baseline} with three components: \textit{Contextualized input encoding} marks the target word within its sentence, \textit{Retrieval} supplies a canonical-form hint from an external datastore, and \textit{Staged decoding} emits the canonical form as an intermediate target before its phonemes (\texttt{<Target>}~$\Gc$~\texttt{<Phoneme>}~$\Pc$).

On \benchmark, all baselines suffer degradation on non-canonical words, with the canonical--non-canonical gap reaching up to 66.8 PER points. Further training on \benchmark{} substantially lowers non-canonical error rates on English and Vietnamese but leaves a large gap. Our compositional baseline consistently reduces non-canonical G2P error relative to matched ByT5 and Qwen2.5-0.5B backbones across all three languages, while the 0.5B variant is also competitive with much larger few-shot frontier LLMs. 

We summarize our contributions as follows:
\begin{itemize}[leftmargin=*, itemsep=2pt, topsep=2pt, parsep=0pt]
\item \textbf{\benchmark}, the first G2P benchmark for user-generated
text across English, Vietnamese, and Korean, with an inference-grounded
taxonomy for fine-grained diagnosis.
\item \textbf{A systematic evaluation} showing how canonical and
non-canonical G2P errors vary across inference types, model families,
and languages.
\item \textbf{A simple compositional baseline} that injects
canonical-form evidence through exact-match lookup and staged decoding,
consistently improving matched supervised backbones on \benchmark.
\end{itemize}

\begin{table*}[t]
    \centering
    \small
    \resizebox{\textwidth}{!}{%
    \begin{tabular}{cll ll ll ll}
    \toprule[2pt]
    & & & \multicolumn{2}{c}{\textbf{English}}
        & \multicolumn{2}{c}{\textbf{Vietnamese}}
        & \multicolumn{2}{c}{\textbf{Korean}} \\
    \cmidrule(lr){4-5}\cmidrule(lr){6-7}\cmidrule(lr){8-9}
    \textbf{Type} & \textbf{ID} & \textbf{Category}
    & $G_{\text{nc}}\!\to\!G_c$ & \textbf{tr+dv / te}
    & $G_{\text{nc}}\!\to\!G_c$ & \textbf{tr+dv / te}
    & $G_{\text{nc}}\!\to\!G_c$ & \textbf{tr+dv / te} \\
    \midrule
    \multirow{3}{*}{\makecell{\textbf{A}\\\textit{Reconstruct}}}
    & A1 & Shortening-V  & pls $\to$ please        & 595/166& \vn{đc} $\to$ \vn{được}   & 3{,}112/327 & \ko{ㅅㄱ} $\to$ \ko{수고}  & 451/67 \\
    & A2 & Shortening-O  & bc $\to$ because & 313/82 & \vn{k} $\to$ \vn{không}   & 4{,}609/496& -- & --   \\
    & A3 & Phrasal (Unp) & idk $\to$ I don't know  & 504/117 & \vn{mn} $\to$ \vn{mọi người} & 2{,}020/225& -- & --   \\ \midrule
    \multirow{2}{*}{\makecell{\textbf{B}\\\textit{Repetition}}}
    & B1 & Lengthening   & soooo $\to$ so          & 175/75 & \vn{điii} $\to$ \vn{đi}   & 475/43 & \ko{구우웃} $\to$ \ko{굿} & 166/66 \\
    & B2 & Iteration & waitwait $\to$ wait wait & 113/44 & \vn{xem xem} $\to$ \vn{xem xem} & 157/38 & \ko{ㅋㅋㅋㅋ} $\to$ \ko{크크크크, 키키키키} & 322/136 \\ \midrule
    \multirow{4}{*}{\makecell{\textbf{C}\\\textit{Pass-through}}}
    & C1 & Eye Direct & luv $\to$ love & 515/140 & \vn{zô} $\to$ \vn{vô}     & 2{,}385/368 & \ko{어케} $\to$ \ko{어케}  & 509/119 \\
    & C2 & Regular & droppin $\to$ dropping  & 1{,}559/405 & \vn{iu} $\to$ \vn{yêu}    & 2{,}164/247 & \ko{걍} $\to$ \ko{걍}      & 175/44 \\
    & C3 & Slang/Teen & tho $\to$ though & 903/236 & \vn{hong} $\to$ \vn{không} & 2{,}897/369 & \ko{막짤} $\to$ \ko{막짤} & 1{,}205/234 \\
    & C4 & Phrasal (Pro) & rofl $\to$ rofl & 166/62 & \vn{GATO} $\to$ \vn{GATO} & 129/39 & \ko{월클} $\to$ \ko{월클} & 89/56 \\
    \bottomrule[2pt]
    \end{tabular}%
    }
    \caption{Diagnostic taxonomy of non-canonical UGT in \benchmark{}, regrouping prior orthographic taxonomies by the inference G2P must perform: three types ordered by inference demand, with nine fine-grained categories. \textbf{(A) Reconstruct}: recover dropped characters. \textbf{(B) Repetition}: collapse repeats. \textbf{(C) Pass-through}: surface already encodes pronunciation. Each cell: example $\Gnc\!\to\!\Gc$ and counts (train+dev / test); ``--'': absent.}
    \label{tab:taxonomy}
\end{table*}
\begin{table}[t]
    \centering
    \small
    \resizebox{0.45\textwidth}{!}{%
    \begin{tabular}{lrrrr}
    \toprule[2pt]
     & \textbf{English} & \textbf{Vietnamese} & \textbf{Korean} & \textbf{Total} \\
    \midrule
    Train  & 3{,}329  &  8{,}597  & 1{,}976  & 13{,}902 \\
    Dev    &    600   &  1{,}129  &    246   &  1{,}975 \\
    Test   & 1{,}100  &  1{,}121  &    275   &  2{,}496 \\
    \midrule
    Total  & 5{,}029  & 10{,}847  & 2{,}497  & 18{,}373 \\
    \bottomrule[2pt]
    \end{tabular}}
    \caption{\benchmark composition by language and split.}
    \label{tab:data}
\end{table}

\section{\benchmark}
\label{sec:data}

We propose \benchmark, the first G2P benchmark for UGT, comprising 18{,}373 sentences across three typologically diverse languages (\Cref{tab:data}): English (5{,}029), Vietnamese (10{,}847), and Korean (2{,}497). We formalize G2P on UGT (\Cref{ssec:task_defs}), group non-canonical words by required inference into a diagnostic taxonomy (\Cref{ssec:taxonomy}), and annotate $(\Gnc, \Gc, \Pc)$ at the target level within full sentential context (\Cref{ssec:data_construction}). This supports aggregate evaluation and category-level diagnosis of G2P failures.

\subsection{Task Formulation}
\label{ssec:task_defs}
\paragraph{Notation.}
Let $S$ denote a sentence and $w_i$ an annotated target token or span in $S$, which is either canonical or non-canonical. In Vietnamese, a single annotated target may span multiple whitespace-delimited units. A canonical word has a spelling listed in the standard dictionary (e.g., \textit{love}, \textit{please}, \vn{\textit{không}}). A non-canonical word $\Gnc$ is a conventionalized variant of a canonical form arising systematically in UGT (e.g., \textit{luv}, \textit{pls}, \vn{\textit{k}}). We denote the canonical form of $\Gnc$ as $\Gc$ and the corresponding gold IPA phoneme sequence as $\Pc$. For each annotated target, we represent the G2P task as a quadruplet $(S, \Gnc, \Gc, \Pc)$, where the non-canonical form $\Gnc$ appears as a token or span in sentence $S$.

\paragraph{Canonical form annotation.}
We adopt the standard formulation of lexical normalization
~\cite{van2018taxonomy, van2021multilexnorm, nguyen2024vilexnorm}:
the mapping $\Gnc \to \Gc$ admits 1-to-1
(\textit{pls} $\to$ \textit{please}), 1-to-N
(\textit{idk} $\to$ \textit{I don't know}), and N-to-1
(\vn{\textit{ch ó}} $\to$ \vn{\textit{chó}}) mappings. We obtain $\Gc$ from LexNorm2015~\cite{baldwin2015shared}, via MultiLexNorm~\cite{van2021multilexnorm}, for English and from ViLexNorm~\cite{nguyen2024vilexnorm} for Vietnamese. For Korean, we draw $\Gnc$ from MultiLexNorm++~\cite{buaphet2026multilexnorm++} and KoMultiText~\cite{choi2023komultitextlargescalekoreantext}, and annotate $\Gc$ from scratch with native-speaker experts. (procedure in \Cref{ssec:appendix-canonical-annotation}).

\paragraph{Task definition.}
Existing G2P benchmarks evaluate word-level conversion in isolation~\cite{bisani2008joint, novak2012wfst, rao2015grapheme}. This is insufficient for UGT: a non-canonical surface like \textit{ig} can normalize to \textit{I guess} or \textit{Instagram} depending on the sentence. \benchmark{} therefore situates every non-canonical word within its original sentence $S$. Given $\Gnc$ and $S$, the task is to predict the gold phoneme sequence $\Pc$ derived from the intended canonical form $\Gc$, which remains latent during evaluation. Evaluation is at the target level for precise error attribution, while conditioning on the full sentence.

\subsection{Diagnostic Taxonomy of Non-Canonical Words}
\label{ssec:taxonomy}
Non-canonical surfaces in UGT vary in the inference required for G2P: some can be read directly, others require recovering dropped characters, and others require collapsing repeated units. Existing UGT taxonomies~\cite{van2018taxonomy} group surfaces by orthographic transformation, which is suited to lexical normalization but not phonemization. \benchmark{} instead regroups these surfaces by the inference required for phonemization (\Cref{tab:taxonomy}), so that failure patterns localize \emph{which kind of inference} a model lacks.

\paragraph{Coarse-grained types.}
We define three coarse types, ordered from the most to the least demanding inference.

\textbf{(A) Reconstruct.} Surface content required for pronunciation has been omitted and must be recovered, possibly as a multi-word expansion (e.g., \textit{idk} $\to$ \textit{I don't know}).

\textbf{(B) Repetition.} Characters or units have been repeated for prosodic emphasis~\cite{brody2011cooooooooooooooollllllllllllll} and must be collapsed (e.g., \textit{soooo} $\to$ \textit{so}).

\textbf{(C) Pass-through.} The surface already preserves the intended pronunciation and can be read directly (e.g., \textit{luv} $\to$ \textit{love}).

\paragraph{Fine-grained categories.}
Within each coarse type, we distinguish 2--4 fine-grained categories reorganized from prior UGT taxonomies~\cite{van2018taxonomy}.
\textbf{Reconstruct} splits by what is omitted (vowels A1, other characters A2, full phrases A3).
\textbf{Repetition} splits by what is repeated (characters B1, units B2).
\textbf{Pass-through} splits by how pronunciation is preserved in the surface (phonetic respelling C1, productive variation C2, lexicalized slang C3, pronounceable acronyms C4).
Examples and per-language counts are shown in \Cref{tab:taxonomy}. Korean Reconstruct collapses to A1, since it consists almost entirely of initial-consonant abbreviations (e.g., \ko{ㄱㅅ} $\to$ \ko{감사}).

\begin{figure*}[t]
    \centering
    \includegraphics[width=\textwidth]{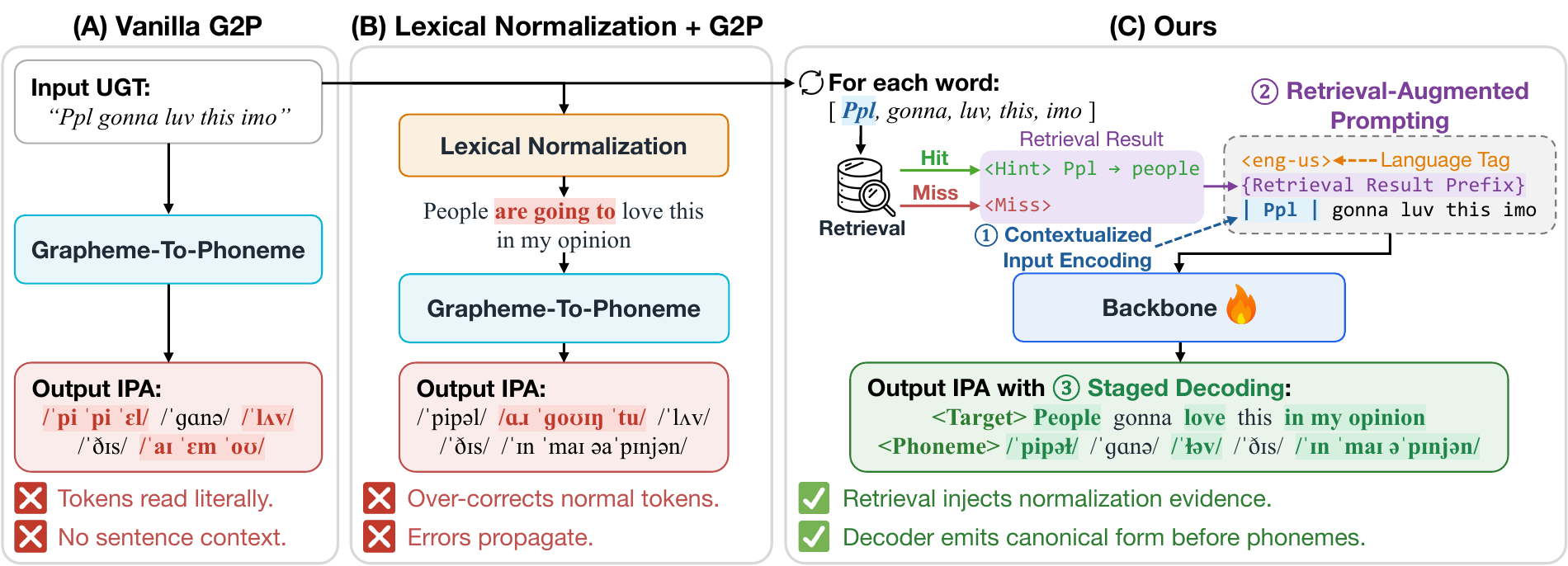}
    \caption{Three approaches to non-canonical G2P. (A) Vanilla G2P reads words literally. (B) A lexical normalization model rewrites the input but overcorrects canonical words, propagating errors downstream. (C) Our retrieval-augmented compositional G2P baseline injects a $\Gnc \rightarrow \Gc$ canonical form hint and emits $\Gc$ as an intermediate before the phonemes.}
    \label{fig:methods}
\end{figure*}

\subsection{Data Construction}
\label{ssec:data_construction}
We construct \benchmark in three stages: source corpora selection, phoneme annotation, and category labeling. We preserve the raw UGT sentence in the released text field. During phoneme construction, non-lexical elements such as punctuation and numbers are dropped or verbalized as appropriate, while unphonemizable items may remain as placeholders. Full details, licenses, and annotation procedures are provided in \Cref{sec:benchmark-construction}.

\paragraph{Source corpora.}
For English, $\Gnc$--$\Gc$ pairs are derived from LexNorm~\cite{baldwin2015shared}, via MultiLexNorm~\cite{van2021multilexnorm}. For Vietnamese, they are derived from ViLexNorm~\cite{nguyen2024vilexnorm}. For Korean, $\Gnc$ is drawn from MultiLexNorm++~\cite{buaphet2026multilexnorm++} and KoMultiText~\cite{choi2023komultitextlargescalekoreantext}, and $\Gc$ is annotated from scratch (\Cref{ssec:appendix-canonical-annotation}). In addition, the English and Vietnamese subsets include 175 and 385 human-curated augmented UGT sentences, respectively, generated from existing UGT patterns. We preserve the natural resource asymmetry across languages rather than down-sampling, ensuring sufficient samples per category in every split. Since the source corpora include UGT from social media, \benchmark inherits some toxic content; we report a sentence-level moderation analysis in \Cref{sec:appendix-moderation}.

\paragraph{Phoneme annotation.}
Target phonemes $\Pc$ are derived from $\Gc$ via IPA-Dict~\cite{ipadict}, an expert-curated grapheme-to-IPA lexicon, and verified by native-speaker experts for language-specific phenomena (English heteronyms, Vietnamese tones, Korean phonological rules). For canonical forms $\Gc$ not covered by the lexicon (e.g., recent loanwords or proper nouns), experts provide $\Pc$ from scratch and cross-verify. Inter-annotator agreement reaches Cohen's $\kappa = 0.86$ (English) and $0.87$ (Vietnamese), both in the almost-perfect range~\cite{landis1977measurement}, with Korean at $0.75$ (substantial) reflecting variability in IPA transcription rather than disagreement on the underlying pronunciation. Per-category agreement and OOV statistics are reported in \Cref{sec:appendix-phoneme-annotation}.

\paragraph{Category labeling.}
Each non-canonical word is assigned to one of the nine categories (\Cref{tab:taxonomy}) through manual annotation by native speakers, with second-pass verification and disagreements reconciled through discussion. Operational definitions and full guidelines are in \Cref{ssec:appendix-category-annotation} (\Cref{tab:category_definitions}).

\section{Method}
\label{sec:method}
Two conventional approaches to non-canonical G2P lack joint supervision of the normalization-phonemization composition (\Cref{fig:methods}). \textbf{Vanilla G2P} (\Cref{fig:methods}~(A)) phonemizes the surface directly, leaving $\Gnc \to \Gc$ unsupervised. \textbf{Lexical normalization + G2P} (\Cref{fig:methods}~(B)) decouples the stages but over-corrects canonical-recoverable surfaces. We use a retrieval-augmented compositional baseline (\Cref{fig:methods}~(C)) that jointly supervises both stages within a single architecture.

Our approach comprises three components. First, \textit{Contextualized input encoding} (\Cref{ssec:method-input}) marks the target within its sentential context. Then, \textit{Retrieval-augmented canonical-form prompting} (\Cref{ssec:method-retrieval}) supplies a $\Gnc \rightarrow \Gc$ hint through an exact-match lookup when the surface form occurs in a training-derived datastore. Finally, \textit{Staged decoding} (\Cref{ssec:method-training}) emits $\Gc$ before $\Pc$, binding the retrieved hint to prediction.

\subsection{Contextualized Input Encoding}
\label{ssec:method-input}
For each annotated target token or span $w_i$ in sentence $S$, we construct the input string by prepending a language tag and surrounding $w_i$ with pipe markers:
\[
\texttt{<lang>}\ w_1\,w_2\,\cdots\,\texttt{|}\,w_i\,\texttt{|}\,\cdots\,w_n.
\]
This format treats $w_i$ as a single annotated target span, even when its surface or canonical form spans multiple whitespace-delimited units (e.g., \textit{idk} $\to$ \textit{I don't know}). Each target is processed as an independent forward pass, so multiple non-canonical words do not accumulate cross-word errors during decoding~\cite{yoon2023mitigating}.

This input format applies uniformly across training and inference, and across different backbone families: ByT5-small~\cite{xue2022byt5} and Qwen 2.5-0.5B~\cite{qwen2025qwen25technicalreport}. Since the \texttt{|} symbol is a regular character in both vocabularies, no special tokens are added. Our approach is therefore backbone-agnostic and does not require backbone-specific tokenization. We use ByT5-small as the default backbone for its byte-level input, which handles \benchmark{}'s diverse scripts without language-specific tokenizers.

\subsection{Retrieval-Augmented Canonical-Form Inference}
\label{ssec:method-retrieval}
Training a model to map $\Gnc$ directly to $\Pc$ provides no explicit supervision for the $\Gnc \to \Gc$ normalization stage: the surface form bears little phonemic correspondence to the target (\textit{idk} vs.\ /\textipa{aI doUnt noU}/). We augment the input with an optional canonical-form hint retrieved through a simple exact-match key--value lookup over a training-derived datastore. 

\paragraph{Retrieval datastore.}
We construct the datastore $\mathcal{D}$ as a key--value map from training-set $(\Gnc, \Gc)$ pairs, with $\Gnc$ as the key and $\Gc$ as the value. For each target $w_i$ in $S$, we retrieve its canonical hint from $\mathcal{D}$:
\begin{equation}
\label{eq:prefix}
  \text{prefix}(w_i) =
  \begin{cases}
    \texttt{<Hint>}\ w_i  \rightarrow \mathcal{D}[w_i] & \text{if } w_i \in \mathcal{D}, \\
    \texttt{<Miss>} & \text{otherwise.}
  \end{cases}
\end{equation}
On a miss, the \texttt{<Miss>} marker explicitly marks retrieval failure as a learnable condition, prompting the model to rely on sentence context for normalization.

\paragraph{Optional LLM-based datastore expansion.}
New non-canonical forms appear continuously, and the training split cannot exhaust the space of variants seen at test time. To extend coverage, we prompt GPT-5.4-mini~\cite{singh2025openai} with seed pairs from $\mathcal{D}$ to generate additional $(\Gnc, \Gc)$ candidates following the same UGT category patterns. Seeds and all filtering are restricted to the training split. We discard candidates whose $\Gnc$ matches a standard word in the train or dev split, to avoid spuriously rewriting canonical inputs, or where $\Gc = \Gnc$, which carries no normalization signal. Prompt details and coverage statistics are provided in \Cref{sec:appendix-db-expansion}.

\paragraph{Hit and Miss prefixes.}
The two special tokens introduced in \Cref{eq:prefix}, \texttt{<Hint>} and \texttt{<Miss>}, are added to the vocabulary. Prepending the retrieval prefix to the contextualized input from \Cref{ssec:method-input} produces the following inputs:

\smallskip
\noindent\textbf{Hit:}\quad
{\small\texttt{<lang> <Hint>}\ $w_i{\to}G_{c,i}$\ \ 
$w_1\,w_2\,\cdots\,$\texttt{|}$w_i$\texttt{|}$\,\cdots\,w_n$}

\smallskip
\noindent\textbf{Miss:}\quad
{\small\texttt{<lang> <Miss>}\ \ 
$w_1\,w_2\,\cdots\,$\texttt{|}$w_i$\texttt{|}$\,\cdots\,w_n$}
\smallskip

The \texttt{<Miss>} marker also distinguishes a non-canonical word absent from $\mathcal{D}$ from a canonical input that requires no retrieval.

\providecommand{\PERipa}{\text{PER}_{\text{ipa}}}

\definecolor{degLarge}{RGB}{180, 25, 25}
\definecolor{degMed}{RGB}{200, 145, 30}  
\definecolor{degSmall}{RGB}{40, 130, 60}

\providecommand{\dlAng}[1]{\textcolor{degLarge}{\scriptsize $\langle$+#1$\rangle$}}
\providecommand{\dmAng}[1]{\textcolor{degMed}{\scriptsize $\langle$+#1$\rangle$}}
\providecommand{\dsAng}[1]{\textcolor{degSmall}{\scriptsize $\langle$+#1$\rangle$}}

\providecommand{\dlarge}[1]{\textcolor{degLarge}{\textbf{#1}}}
\providecommand{\dmed}[1]{\textcolor{degMed}{\textbf{#1}}}
\providecommand{\dgood}[1]{\textcolor{degSmall}{\textbf{#1}}}

\begin{table*}[t]
  \centering
  \small
  \setlength{\tabcolsep}{3.5pt}
  \resizebox{\textwidth}{!}{%
  \begin{tabular}{lrrlcrrlcrrl}
    \toprule[2pt]
    & \multicolumn{3}{c}{\textbf{English}} &
    & \multicolumn{3}{c}{\textbf{Vietnamese}} &
    & \multicolumn{3}{c}{\textbf{Korean}} \\
    \cmidrule(lr){2-4} \cmidrule(lr){6-8} \cmidrule(lr){10-12}
    Model
      & $\PERipa$\,\dn & $\PERc$\,\dn & $\PERnc\,\langle\Delta\rangle$\,\dn &
      & $\PERipa$\,\dn & $\PERc$\,\dn & $\PERnc\,\langle\Delta\rangle$\,\dn &
      & $\PERipa$\,\dn & $\PERc$\,\dn & $\PERnc\,\langle\Delta\rangle$\,\dn \\
    \midrule

    \multicolumn{12}{l}{\textcolor{cGreen}{\textit{Rule-based G2P}}} \\
    Epitran\,{\scriptsize(--)}
      & 33.4 & 13.4 & 45.1\,\dlAng{31.7} &
      & 32.7 & 34.7 & 75.1\,\dlAng{40.4} &
      & 47.0 & 37.5 & 50.7\,\dmAng{13.2} \\
    viG2P\,{\scriptsize(--)}
      &  --  &  --  &  --  &
      & 25.4 & 27.3 & 78.4\,\dlAng{51.1} &
      &  --  &  --  &  --  \\
    \addlinespace[2pt]

    \multicolumn{12}{l}{\textcolor{cGreen}{\textit{Supervised G2P {\normalfont(trained on IPA-Dict canonical lexicon)}}}} \\
    Transformer~\citeyearpar{yolchuyeva2020transformer}\,{\scriptsize(2.0M)}
      & 17.4 & 12.9 & 44.8\,\dlAng{31.9} &
      & 15.3 &  5.8 & 68.4\,\dlAng{62.6} &
      &  7.5 &  9.5 & 54.8\,\dlAng{45.3} \\
    ByT5-tiny~\citeyearpar{byt5g2p}\,{\scriptsize(7.3M)}
      & 17.0 & 18.8 & 46.4\,\dmAng{27.6} &
      & \textbf{15.2} &  6.0 & 71.9\,\dlAng{65.9} &
      &  7.2 &  6.8 & 51.1\,\dlAng{44.3} \\
    GBERT~\citeyearpar{dong2022neural}\,{\scriptsize(8.0M)}
      & 21.5 & 17.6 & 45.0\,\dmAng{27.4} &
      & 24.0 & 30.8 & 93.0\,\dlAng{62.2} &
      & 11.7 & 16.8 & 45.9\,\dmAng{29.1} \\
    ByT5~\citeyearpar{xue2022byt5}\,{\scriptsize(300M)}
      & \textbf{13.5} & 10.5 & 46.6\,\dlAng{36.1} &
      & \textbf{15.2} &  5.9 & 72.7\,\dlAng{66.8} &
      &  \textbf{6.7} &  4.2 & 46.2\,\dlAng{42.0} \\

    \multicolumn{12}{l}{\textcolor{cGreen}{\textit{Supervised G2P {\normalfont(further trained on \benchmark)}}}} \\
    Transformer\,{\scriptsize(2.0M)}
      & 18.8 & 10.3 & 32.7\,\dmAng{22.4} &
      & 17.8 &  5.9 & 33.7\,\dmAng{27.8} &
      &  7.5 &  5.8 & 38.7\,\dlAng{32.9} \\
    ByT5-tiny\,{\scriptsize(7.3M)}
      & 30.1 &  9.7 & 30.8\,\dmAng{21.1} &
      & 17.0 &  7.7 & 43.6\,\dlAng{35.9} &
      &  8.1 & \textbf{4.0} & 39.0\,\dlAng{35.0} \\
     GBERT\,{\scriptsize(8.0M)}
        & 25.6 &  7.8 & 50.5\,\dlAng{42.7} &
        & 50.3 &  5.6 & 60.4\,\dlAng{54.8} &
        &  8.5 &  5.8 & \textbf{31.6}\,\dmAng{25.8} \\
    ByT5\,{\scriptsize(300M)}
      & 15.9 &  5.2 & 29.8\,\dmAng{24.6} &
      & 17.3 &  7.7 & 43.2\,\dlAng{35.5} &
      &  7.2 &  4.3 & 47.5\,\dlAng{43.2} \\
    \midrule
    \rowcolor{Blue}
    \textbf{Ours} (ByT5)\,{\scriptsize(300M)}
      & 17.8 & \textbf{4.7} & 22.5\,\dmAng{17.8} &
      & 15.9 & 5.9 & 26.7\,\dmAng{20.8} &
      &  9.0 & 5.1 & 32.0\,\dmAng{26.9} \\
    \rowcolor{Blue}
    \textbf{Ours} (Qwen 2.5-0.5B)\,{\scriptsize(494M)}
      & 19.5 & 10.2 & \textbf{14.3}\,\dsAng{4.1} &
      & 15.5 & \textbf{4.0} & \textbf{20.0}\,\dmAng{16.0} &
      &  8.2 & 6.0 & 32.4\,\dmAng{26.4} \\
    \bottomrule[2pt]
  \end{tabular}%
  }
   \caption{Comparison with existing G2P models on \benchmark. PER (\%) on canonical ($\PERc$) and non-canonical ($\PERnc$) words, with the within-input gap $\Delta = \PERnc - \PERc$. Color encodes $\Delta$: \dlarge{$\geq 30$\,pp} (large gap), \dmed{$10$--$30$\,pp} (medium), \dgood{$<10$\,pp} (small). Bold: best per column.}
  \label{tab:main_ipa}
\end{table*}

\subsection{Staged Decoding}
\label{ssec:method-training}
To bind the retrieved hint to prediction, we make the canonical form an explicit intermediate target during decoding. The model produces the normalization output $\Gc$ before the phoneme sequence $\Pc$ in a single autoregressive trajectory, with the target sequence structured as follows:
\begin{align*}
\text{Non-canonical:} \quad & 
\underbrace{\texttt{<Target>}\,\Gc}_{\text{normalization}}\,
\underbrace{\texttt{<Phoneme>}\,\Pc}_{\text{phonemization}}, 
\\
\text{Canonical:} \quad & \texttt{<Phoneme>}\,\Pc.
\end{align*}
For canonical words, the model bypasses the normalization stage and emits $\Pc$ directly, aligning with conventional G2P training. For non-canonical words, it generates $\Gc$ as a learned intermediate before $\Pc$, providing explicit supervision for both the normalization stage (via $\Gc$) and the phonemization stage (via $\Pc$). The model is trained end-to-end with cross-entropy loss over the full target sequence, without auxiliary objectives or additional loss terms.

Together, the three components make canonical-form modeling explicit within a single G2P architecture, supervising normalization and phonemization without sacrificing canonical-token accuracy.

\section{Experiments}
\label{sec:experiments}

Using the taxonomy of \Cref{ssec:taxonomy}, we quantify the  canonical-to-non-canonical gap on \benchmark, decompose it by category, and compare our model against two alternatives: a normalize-then-phonemize pipeline and frontier LLMs.

\subsection{Experimental Setup}
\label{ssec:setup}

\paragraph{Evaluation protocol.}
Models take the full sentence as input and predict the marked target word. We report Phoneme Error Rate (PER, \%) on \benchmark's test split as $\PERc$ (canonical) and $\PERnc$ (non-canonical), with $\Delta = \PERnc - \PERc$. For lexicon-only supervised baselines, we additionally report $\PERipa$ on the IPA-dict~\cite{ipadict} lexicon to track canonical coverage.

\paragraph{Baselines.}
We compare our model against task-relevant G2P, normalization, and frontier-LLM baselines.
\textit{Rule-based G2P}: Epitran~\cite{mortensen2018epitran} and viG2P~\cite{vig2p}.
\textit{Supervised G2P}: Transformer~\cite{yolchuyeva2020transformer}, ByT5/ByT5-tiny~\cite{byt5g2p}, and GBERT~\cite{dong2022neural}, each evaluated in two training regimes: \emph{lexicon-only} (trained on canonical IPA-Dict) and \emph{fine-tuned} (further trained on \benchmark).
Frontier LLMs: GPT-4o, GPT-5.4-mini~\citep{singh2025openai},
Gemini-2.5-flash, Gemini-3-flash, and Claude-Sonnet-4.6,
evaluated under the nominal $K=9$ in-context setting described in
\Cref{sec:appendix-prompts},
following prior G2P-with-LLM work~\citep{han2024improving}.
\textit{Task-relevant baselines}: ICKR~\cite{han2024improving}, R-G2P~\cite{zhao2022r}, and lexical-normalization $+$ G2P pipelines using MoNoise~\citep{van2019monoise} or PolyNorm~\citep{wong2025polynorm}. The full cross-lingual comparison is reported in \Cref{tab:appendix-task-baselines}.

\paragraph{Implementation.}
We use ByT5-small as the default backbone, with Qwen2.5-0.5B for backbone ablation.
Training uses AdamW ($\mathrm{lr} = 10^{-4}$, batch size=32, 10 epochs) on 4 NVIDIA V100 GPUs.
The retrieval datastore $\mathcal{D}$ uses the exact-match lookup described in
\Cref{ssec:method-retrieval}.
For the frontier LLMs, we query official APIs under the nominal $K=9$
in-context setting described in \Cref{sec:appendix-prompts}.
Model identifiers and access dates are listed in \Cref{sec:appendix-training};
the prompt template is provided in \Cref{sec:appendix-prompts}.

\definecolor{impLarge}{RGB}{0, 120, 40}
\definecolor{impMed}{RGB}{60, 160, 70}
\definecolor{impSmall}{RGB}{90, 170, 130}
\definecolor{regLarge}{RGB}{180, 20, 20}
\definecolor{regMed}{RGB}{220, 80, 60}
\definecolor{regSmall}{RGB}{210, 120, 100}

\newcommand{\aimproveLG}[1]{\textcolor{impLarge}{\footnotesize \,($-$#1)}}
\newcommand{\aimproveMD}[1]{\textcolor{impMed}{\footnotesize \,($-$#1)}}
\newcommand{\aimproveSM}[1]{\textcolor{impSmall}{\footnotesize \,($-$#1)}}
\newcommand{\aregressLG}[1]{\textcolor{regLarge}{\footnotesize \,($+$#1)}}
\newcommand{\aregressMD}[1]{\textcolor{regMed}{\footnotesize \,($+$#1)}}
\newcommand{\aregressSM}[1]{\textcolor{regSmall}{\footnotesize \,($+$#1)}}
\newcommand{\aneutral}{\textcolor{gray}{\footnotesize \,($\pm$0.0)}}

\newcommand{\cmark}{\textcolor{impLarge}{\ding{51}}}
\newcommand{\xmark}{\textcolor{regLarge}{\ding{55}}}

\begin{table*}[t]
    \centering
    \resizebox{\textwidth}{!}{%
        \begin{tabular}{c|cc|ccc|ccc|ccc}
            \toprule[2pt]
            \multirow{2}{*}{\textbf{Input}} & 
            \multirow{2}{*}{\makecell{\textbf{Retrieval}\\\textbf{Augmentation}}} &
            \multirow{2}{*}{\makecell{\textbf{Staged}\\\textbf{Decoding}}}
            & \multicolumn{3}{c|}{\textbf{English}}
            & \multicolumn{3}{c|}{\textbf{Vietnamese}}
            & \multicolumn{3}{c}{\textbf{Korean}} \\
             & & & $\PERo$\,\dn & $\PERc$\,\dn & $\PERnc$\,\dn & $\PERo$\,\dn & $\PERc$\,\dn & $\PERnc$\,\dn & $\PERo$\,\dn & $\PERc$\,\dn & $\PERnc$\,\dn \\
            \midrule
            Word$\to$Word  & \xmark & \xmark & 7.9 & 5.2 & 29.8 & 14.1 & 7.7 & 43.2 & 17.8 & 4.3 & 47.5 \\
            Sent$\to$Sent  & \xmark & \xmark
            & 19.7\,\aregressLG{11.8} & 18.0\,\aregressLG{12.8} & 30.5\,\aregressSM{0.7}
            & 29.7\,\aregressLG{15.6} & 19.2\,\aregressLG{11.5} & 70.7\,\aregressLG{27.5}
            & 33.5\,\aregressLG{15.7} & 41.6\,\aregressLG{37.3} & 31.3\,\aimproveLG{16.2}\\
            \midrule
            Sent$\to$Word & \xmark & \xmark
            & 7.6\,\aimproveSM{0.3} & 5.4\,\aregressSM{0.2} & 25.5\,\aimproveLG{4.3}
            & 15.0\,\aregressSM{0.9} & 9.1\,\aregressSM{1.4} & 41.9\,\aimproveSM{1.3}
            & \textbf{12.0}\,\aimproveMD{5.8} & 4.7\,\aregressSM{0.4} & 28.1\,\aimproveLG{19.4} \\
            Sent$\to$Word & \xmark & \cmark
            & 8.9\,\aregressSM{1.0} & 4.9\,\aimproveSM{0.3} & 30.1\,\aregressSM{0.3}
            & 10.1\,\aimproveMD{4.0} & 5.9\,\aimproveSM{1.8} & 31.8\,\aimproveLG{11.4}
            & 13.1\,\aimproveMD{4.7} & \textbf{4.0}\,\aimproveSM{0.3} & 33.1\,\aimproveLG{14.4} \\
            Sent$\to$Word & \cmark & \xmark
            & 7.0\,\aimproveSM{0.9} & 5.0\,\aimproveSM{0.2} & 22.5\,\aimproveLG{7.3}
            & 9.6\,\aimproveMD{4.5} & 5.8\,\aimproveSM{1.9} & 29.1\,\aimproveLG{14.1}
            & 12.4\,\aimproveMD{5.4} & 5.5\,\aregressSM{1.2} & \textbf{27.7}\,\aimproveLG{19.8} \\
            \rowcolor{Blue}
            Sent$\to$Word & \cmark & \cmark
            & \textbf{6.8}\,\aimproveMD{1.1} & \textbf{4.7}\,\aimproveSM{0.5} & \textbf{22.5}\,\aimproveLG{7.3}
            & \textbf{9.3}\,\aimproveMD{4.8} & \textbf{5.9}\,\aimproveSM{1.8} & \textbf{26.7}\,\aimproveLG{16.5}
            & 13.5\,\aimproveMD{4.3} & 5.1\,\aregressSM{0.8} & 32.0\,\aimproveLG{15.5} \\
            \bottomrule[2pt]
        \end{tabular}}
    \caption{Component ablation of our retrieval-augmented G2P model on ByT5-small. 
    \textcolor{impLarge}{Green}/\textcolor{regLarge}{Red}: gain/regression vs.\ the \textit{Word$\to$Word} baseline. 
    \textbf{Retrieval Aug}: $G_{\text{nc}}{\rightarrow}G_c$ hints from the datastore. \textbf{Staged Decoding}: \texttt{<Target>}\,$\rightarrow$\,\texttt{<Phoneme>} emission. 
    Bold: column-best per language (PER, \%).}
    \label{tab:component_ablation}
\end{table*}

\subsection{Main Results}
\label{ssec:main_results}

\paragraph{All baselines exhibit large canonical-to-non-canonical gaps.}
\Cref{tab:main_ipa} reports PER for rule-based and supervised baselines on \benchmark. The within-input gap $\Delta$ ranges from $+27.4$ to $+36.1$ on English, $+40.4$ to $+66.8$ on Vietnamese, and $+13.2$ to $+45.3$ on Korean. The pattern persists across rule-based~\cite{mortensen2018epitran}, transformer~\cite{yolchuyeva2020transformer}, and byte-level~\cite{byt5g2p} architectures, and is most severe on Vietnamese, where supervised baselines average $\Delta \approx +64$ despite $\PERc$ below $8\%$.

\paragraph{Further training on \benchmark reduces $\PERnc$ but does not close $\Delta$.}
Training the supervised baselines on \benchmark's training split substantially lowers $\PERnc$ on English ($46.6 \to 29.8$) and Vietnamese ($72.7 \to 43.2$), but the strongest fine-tuned model (ByT5) still shows $\Delta = +24.6$ on English and $+35.5$ on Vietnamese. Korean even slightly worsens ($46.2 \to 47.5$), reflecting its Pass-through--dominated composition in \Cref{tab:taxonomy}, where the surface already encodes pronunciation and further fine-tuning provides little signal. The remaining $\Delta$ shows that further training alone does not eliminate the non-canonical gap, motivating explicit surface-to-canonical modeling.

\paragraph{Explicit canonical-form modeling improves matched backbones.}
Relative to matched ByT5 fine-tuning, our formulation lowers $\PERnc$ from $29.8$ to $22.5$ on English, $43.2$ to $26.7$ on Vietnamese, and $47.5$ to $32.0$ on Korean. The same pattern holds for Qwen2.5-0.5B: $14.8\to14.3$, $23.1\to20.0$, and $77.3\to32.4$, respectively. Three-seed runs show stable English/Vietnamese results, while Korean has higher variance; paired bootstrap tests on the reported checkpoints give confidence intervals excluding zero for all six matched-backbone $\PERnc$ comparisons (\Cref{tab:appendix-stability,tab:appendix-bootstrap}). The largest matched-backbone reductions occur on Vietnamese and Korean, while English Qwen improves from $14.8$ to $14.3$ $\PERnc$.

\subsection{Analysis}
\label{sec:experiments-analysis}

\paragraph{Sentence context helps as evidence, not as a generation target.}
\Cref{tab:component_ablation} shows that generating the full sentence (Sent$\to$Sent) hurts $\PERc$ across all three languages, reflecting exposure bias from long autoregressive outputs~\cite{yoon2023mitigating} (Korean $\PERc$ rises by $37.3$ points). Sent$\to$Word avoids this trade-off and lowers Korean $\PERnc$ ($47.5 \to 28.1$), since Korean is dominated by pass-through forms that sentence context alone can disambiguate. Vietnamese, by contrast, improves little under Sent$\to$Word, motivating the explicit canonical-form evidence that the lookup provides.

\paragraph{Exact-match lookup supplies canonical-form evidence; staged decoding encourages the model to use it.}
On top of Sent$\to$Word, retrieval alone reduces $\PERnc$ from $41.9$ to $29.1$ on Vietnamese and $25.5$ to $22.5$ on English, supplying the surface-to-canonical evidence the surface lacks. Staged decoding alone improves Vietnamese ($41.9 \to 31.8$) but worsens English ($25.5 \to 30.1$) and Korean ($28.1 \to 33.1$): without a retrieval hint, forcing a canonical intermediate invites hallucinated reconstructions. The two together attain the lowest $\PERnc$ on English ($22.5$) and Vietnamese ($26.7$) while keeping $\PERc$ at or below the no-retrieval level. Korean shows no further gain, since its bottleneck has shifted from normalization to phonemization.

\begin{figure}[t]
\begin{center}
\includegraphics[width=\linewidth]{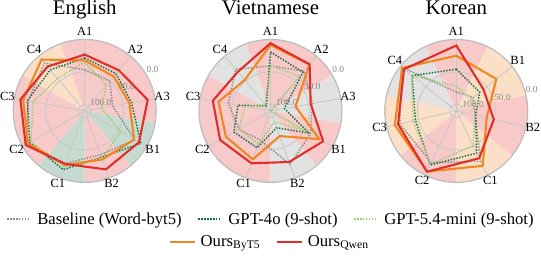}
\end{center}
\caption{$\PERnc$ (\%) by fine-grained category. Gains concentrate on \emph{Reconstruct} (A), with smaller margins on \emph{Repetition} (B) and \emph{Pass-through} (C).}
\label{fig:per-category}
\end{figure}

\begin{table}[t]
\centering
\small
\setlength{\tabcolsep}{8pt}
\begin{tabular}{l cc}
\toprule[2pt]
& \multicolumn{2}{c}{\textbf{English}} \\
\cmidrule(lr){2-3}
Model & $\PERc$ & $\PERnc$ \\
\midrule
word-ByT5 (no norm.)                              &  5.2          & 29.8          \\
\midrule
word-ByT5 + MoNoise~\citeyearpar{van2019monoise}  & 14.2          & 18.0          \\
word-ByT5 + GPT-5.4-mini                          & 13.6          & 18.6          \\
\midrule
\rowcolor{Blue} \textbf{Ours}\,(ByT5)             & \textbf{4.7}  & 22.5          \\
\rowcolor{Blue} \textbf{Ours}\,(Qwen 2.5-0.5B)             & 10.2          & \textbf{14.3} \\
\bottomrule[2pt]
\end{tabular}
\caption{Comparison with two-stage LexNorm $+$ G2P pipelines on English (PER, \%).}
\label{tab:lexnorm_pipeline}
\end{table}

\paragraph{Improvements track the inference demand of each category.}
\Cref{fig:per-category} decomposes $\PERnc$ along the taxonomy of \Cref{tab:taxonomy}, which predicts where canonical-form modeling should help and where it should add little. The results match.

Reconstruct (A) is where the surface and canonical form share the least phonemic evidence (e.g., \textit{idk}\,$\to$\,\textit{I don't know}), so our model's gain is largest: with the Qwen 2.5 (0.5B) backbone, A-average $\PERnc$ drops by $33.4$ points on English ($54.4 \to 21.0$) and $15.2$ on Vietnamese ($38.3 \to 23.1$). On English A3 specifically, our model reaches $11.5$, compared with $66.3$ for the word-level baseline and $36.8$ for GPT-4o, yielding the largest separation on a category that requires phrasal canonical-form reconstruction.

Repetition (B) is intermediate: the surface encodes the pronunciation but with redundant content that must be collapsed. Our model attains the best $\PERnc$ on English B2 and sharply reduces Vietnamese B2 against frontier LLMs, though the word-level ByT5 baseline remains marginally stronger on Vietnamese B2 ($22.5$ vs.\ $26.3$), since the redundant surface still preserves the underlying phonemes.

Pass-through (C) is mixed: the surface already encodes the pronunciation, so the canonical-form intermediate is no longer strictly necessary. Where the baseline is near-zero (Korean C4, $1.9$), our model offers no benefit; where the surface remains non-trivial to phonemize, the gain returns. Our model attains the best $\PERnc$ on nine of twelve Pass-through subcategories, so the cost-benefit depends on residual phonemization difficulty rather than the coarse category label.

\begin{figure}[t]
\begin{center}
\includegraphics[width=\linewidth]{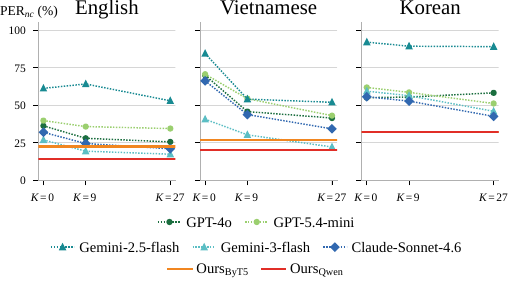}
\end{center}
\caption{Shot scaling for frontier LLMs on \benchmark.
$\PERnc$ (\%) under the nominal $K \in \{0, 9, 27\}$ shot settings; Korean uses 0, 7, and 21 actual in-context examples because A2 and A3 are absent.}
\label{fig:shot_scaling}
\end{figure}

\paragraph{Our approach outperforms two-stage pipelines without trading canonical for non-canonical accuracy.} 
\Cref{tab:lexnorm_pipeline} compares our model against two-stage English pipelines as shown in \Cref{fig:methods} (B) that pair a lexical normalizer with word-level ByT5: MoNoise~\cite{van2019monoise} and GPT-5.4-mini~\cite{singh2025openai}. English is the only language where this comparison is possible, since dedicated lexical normalization models exist for English but not for the lower-resource languages we evaluate. The two pipelines reduce $\PERnc$ from $29.8$ to roughly $18$, but $\PERc$ degrades from $5.2$ to over $13$. Our model avoids this trade-off: with the ByT5 backbone, $\PERc$ stays at $4.7$ and $\PERnc$ drops to $22.5$; with Qwen 2.5-0.5B, $\PERnc$ reaches the lowest value across all systems ($14.3$) while $\PERc$ ($10.2$) remains well below the pipelines.

\paragraph{Our approach remains competitive with frontier LLMs.}
\Cref{fig:shot_scaling} compares our approach against five frontier LLMs under the nominal $K \in \{0, 9, 27\}$ shot settings. Across all three languages, our 0.5B model achieves performance competitive with or better than the evaluated frontier LLMs. On English, the strongest frontier model (Gemini-3-flash) reaches $\PERnc = 17.4$ under $K=27$, while our Qwen2.5-0.5B model achieves $14.3$. On Vietnamese, frontier LLMs exhibit substantially larger errors on several UGT categories, including GPT-5.4-mini on Reconstruct A3 ($\PERnc = 109.6$; \Cref{fig:per-category}), whereas our model achieves the lowest overall $\PERnc$ of $20.0$. On Korean, our ByT5 and Qwen2.5-0.5B variants achieve $\PERnc$ of $32.0$ and $32.4$, respectively, compared with $42.6$ for the strongest evaluated frontier baseline (Claude-Sonnet-4.6 under $K=27$). These results further highlight the difficulty of canonical-form inference for UGT, particularly in Vietnamese and Korean.

\paragraph{The training-set datastore alone closes most of the gap to oracle retrieval.}
\Cref{tab:datastore} compares four datastores: no retrieval, training-set entries only, training-set with LLM-generated augmentation, and an oracle that uses gold canonical forms at test time. The training-set datastore alone reaches $\PERnc = 22.7$ on English, within $2.1$ points of the oracle ($20.6$), and $26.7$ on Vietnamese, within $0.8$ points of the oracle ($25.9$). Adding LLM-generated $(\Gnc, \Gc)$ pairs reduces English $\PERnc$ by only $0.2$ point and leaves Vietnamese unchanged. The deployable datastore approaches oracle performance without test-time canonical-form access.

\begin{table}[t]
\centering
\small
\setlength{\tabcolsep}{8pt}
\begin{tabular}{l ccc}
\toprule[2pt]
\multirow{2}{*}{\textbf{Datastore} $\mathcal{D}$}
  & \multicolumn{3}{c}{$\PERnc$ (\%)} \\
\cmidrule(lr){2-4}
  & EN & VI & KO \\
\midrule
No retrieval ($\emptyset$)        & 30.1          & 31.8          & 33.1 \\
Train-only                        & 22.7          & 26.7          & 32.4 \\
\rowcolor{Blue}
\quad + LLM Aug                   & \textbf{22.5} & \textbf{26.7} & \textbf{32.0}    \\
\midrule
\textit{Oracle} (not deployable)  & 20.6          & 25.9          & 18.6   \\
\bottomrule[2pt]
\end{tabular}
\caption{Retrieval datastore $\mathcal{D}$ sensitivity (ByT5). \textit{Oracle} uses gold canonical forms (not deployable).}
\label{tab:datastore}
\end{table}

\paragraph{Downstream TTS-ASR validation.}
On downstream TTS-ASR evaluations with eSpeak NG and Whisper large-v3~\citep{radford2023robust} (\Cref{sec:appendix-tts}), our ByT5 variant attains the lowest WER on English ($0.359$), while our Qwen2.5-0.5B variant attains the lowest observed WER among predicted-input systems on Vietnamese ($0.640$). These results are consistent with the token-level PER trends.
\section{Conclusion}
\label{sec:conclusion}
We diagnose user-generated text as a setting where G2P must compose surface-to-canonical inference with phonemization, and introduce \benchmark, the first G2P benchmark for UGT with an inference-grounded taxonomy across English, Vietnamese, and Korean. Our retrieval-augmented compositional baseline makes canonical-form inference explicit through exact-match hints and staged decoding, reducing non-canonical PER relative to matched ByT5 and Qwen2.5-0.5B backbones across all three languages. The gains are largest on Reconstruct cases, while the 0.5B variant remains competitive with the evaluated frontier LLMs. Together, \benchmark provides a multilingual benchmark and diagnostic framework for studying canonical-form inference in UGT phonemization.

\section*{Limitations}
We include English and Vietnamese TTS-ASR evaluations (\Cref{sec:appendix-tts}) showing
that lower PER is accompanied by lower word error rate for several systems, but we
do not report subjective MOS ratings or evaluate across multiple TTS systems;
characterizing the full relationship between PER and TTS quality on UGT is
left to future work. \benchmark spans three typologically diverse languages
(alphabetic English, tonal-diacritic Vietnamese, syllabic-block Korean) but
does not cover logographic, abjad, or abugida writing systems, where UGT
phenomena and the relevant normalization signals may differ. Whether the
compositional decomposition transfers to these scripts is an open question.
Our retrieval datastore is built from training-set pairs and an
LLM-based expansion. UGT vocabulary drifts continuously, so maintaining
coverage on emerging non-canonical forms will require periodic datastore
refresh; we leave the dynamics of such refresh, and zero-shot extensions of
the retrieval mechanism, to future work. Beyond these scope limitations,
more accurate UGT phonemization is dual-use: it improves TTS intelligibility
on harmful UGT as well as benign UGT, and \benchmark inherits toxic content
from its source corpora, whose moderation profile we report in
\Cref{sec:appendix-moderation}.

\section*{Acknowledgments}

This work was partly supported by NAVER Cloud Corporation, 
the Institute of Information \& Communications Technology Planning \& Evaluation (IITP) 
grant funded by the Korean government (MSIT) 
(No. RS-2020-II201373, Artificial Intelligence Graduate School Program, Hanyang University), 
and the Institute of Information \& Communications Technology Planning \& Evaluation (IITP) 
grant funded by the Korean government (MSIT) 
(No. RS-2025-02215122, Development and Demonstration of Lightweight AI Model for Smart Homes).

\bibliography{custom}

\appendix

\crefalias{section}{appendix}
\crefalias{subsection}{appendix}
\crefalias{subsubsection}{appendix}
\crefname{appendix}{appendix}{appendices}
\Crefname{appendix}{Appendix}{Appendices}

\clearpage

\section{Creating \benchmark}
\label{sec:benchmark-construction}

This appendix details the dataset construction procedure summarized in \Cref{ssec:data_construction}, including the procedures for source corpus collection and preprocessing, as well as the annotation pipelines for canonical forms, categories, and phonemes.

\subsection{Source Corpora and Preprocessing}

\paragraph{Source corpora.}
Non-canonical tokens $G_{\mathrm{nc}}$ in \benchmark are drawn from existing user-generated text resources, with small human-curated augmented subsets for English and Vietnamese.
For English, we use MultiLexNorm~\citep{van2021multilexnorm}, whose English data builds on LexNorm2015~\citep{baldwin2015shared}. For Vietnamese, we use ViLexNorm~\citep{nguyen2024vilexnorm}, a lexical normalization corpus of Vietnamese social media text whose
annotations align with our $G_{\mathrm{nc}} \rightarrow G_{\mathrm{c}}$
formulation. For Korean, we source $G_{\mathrm{nc}}$ from MultiLexNorm++~\citep{buaphet2026multilexnorm++}, supplemented by the Korean UGT corpus of \citet{choi2023komultitextlargescalekoreantext}, and re-annotate $G_{\mathrm{c}}$ from scratch using the protocol in \Cref{ssec:appendix-canonical-annotation}. In addition, the English and Vietnamese subsets include 175 and 385 human-curated augmented UGT samples, respectively, generated from existing UGT patterns.

\paragraph{Licenses.}
Source corpora are used under their original licensing terms. MultiLexNorm's English data builds on LexNorm2015~\citep{baldwin2015shared}, which has no explicit license; following the convention adopted by MultiLexNorm~\citep{van2021multilexnorm}, attribution is provided through citation. The English subset of \benchmark is provided strictly for academic and research purposes, and commercial use is prohibited. ViLexNorm~\citep{nguyen2024vilexnorm} is released under CC BY-NC-SA~4.0. The Korean subset is derived from KoMultiText~\citep{choi2023komultitextlargescalekoreantext}, partly via MultiLexNorm++~\citep{buaphet2026multilexnorm++}, and is released under
Apache 2.0. \benchmark is released with per-language licenses reflecting these
upstream terms.

\paragraph{Data provenance and consent.}
\benchmark is constructed from previously released UGT datasets rather than through direct collection from the original social-media users.
We therefore do not separately recruit or obtain consent from the original text authors, and instead follow the data-use conditions and licensing terms of the respective source datasets.

\paragraph{Preprocessing.}
We preserve the original sentence boundaries, casing, and surrounding context of each $G_{\mathrm{nc}}$ token in the UGT sentence. Non-lexical elements are handled during phoneme construction rather than uniformly removed from the raw text: punctuation may be dropped, numbers may be verbalized, and elements that cannot be phonemized may remain as placeholders.

\subsection{Canonical Form Annotation}
\label{ssec:appendix-canonical-annotation}

For English and Vietnamese, canonical forms $G_c$ are obtained directly from the source lexical normalization corpora~\citep{baldwin2015shared, nguyen2024vilexnorm}, which provide surface-to-canonical mappings.

\paragraph{Manual annotation for Korean.}
Unlike English and Vietnamese, no comparably large-scale lexical normalization corpus exists for Korean.
We therefore annotate $G_c$ from scratch with two native-speaker experts following a category-specific protocol that reflects the structure of Korean UGT; the full annotation guide is provided below:

\begin{itemize}[leftmargin=*, itemsep=0pt, topsep=3pt]
\item \textbf{A1 (Shortening-V, \ko{초성체})}: Consonant abbreviations are expanded to their full Hangul form (e.g., \ko{ㄱㅅ} $\to$ \ko{감사}, \ko{ㅇㄱㄹㅇ} $\to$ \ko{이거레알}).%
\item \textbf{B1 (Lengthening)}: The elongation of vowel or syllable is collapsed to their base form (e.g., \ko{꺄아악} $\to$ \ko{꺅}).
\item \textbf{B2 (Repetition)}: For consonant-only repetition (e.g., \ko{ㅋㅋㅋㅋ}), $G_c$ is the Hangul realization preserving the intended count (e.g., \ko{크크크크}, \ko{키키키키}, \ko{킥킥킥킥}; recorded as comma-separated alternatives when more than one is acceptable). For syllable- or word-level repetition (e.g., \ko{빨리빨리}, \ko{감사감사}), the surface itself is the intended pronunciation, so $G_c$ retains the 
repetition.
\item \textbf{C1--C4 (Pass-through)}: The surface form already encodes the intended pronunciation, so $G_c = \Gnc$ (e.g., \ko{머해}, \ko{킹받다}, \ko{갑분싸}).
\end{itemize}
Tokens from the source corpora that turn out to be standard Korean rather than UGT are removed from the benchmark.

\subsection{Category Annotation}
\label{ssec:appendix-category-annotation}

\begin{table*}[h]
    \centering
    \small
    \begin{tblr}{
        colspec = {Q[c,m,1.7cm] Q[c,m,0.5cm] Q[l,m,2.5cm] Q[l,m,6.3cm] Q[l,m,2.5cm]},
        row{1} = {font=\bfseries},
        rowsep = 2pt,
      }
    \toprule[2pt]
    Type & ID & Category & Description & Examples \\
    \midrule
    \SetCell[r=3]{c,m} {\textbf{A}\\ \textit{Reconstruct}}
    & A1 & {Shortening-V\\ (Vowel Drop)}
    & Canonical-form consonants are preserved, vowels are dropped.
    & {\textit{pls}\,$\to$\,\textit{please}\\ \vn{\textit{đc}}\,$\to$\,\vn{\textit{được}}} \\
    & A2 & {Shortening-O\\ (Other Drop)}
    & One or more characters dropped from a single canonical word, beyond vowel-only deletion.
    & {\textit{bc}\,$\to$\,\textit{because}\\ \vn{\textit{k}}\,$\to$\,\vn{\textit{không}}} \\
    & A3 & {Phrasal\\ (Unpronounceable)}
    & A multi-word canonical expansion whose pronunciation is not directly recoverable from the surface form.
    & {\textit{idk}\,$\to$\,\textit{I don't know}\\ \vn{\textit{mn}}\,$\to$\,\vn{\textit{mọi người}}} \\
    \midrule
    \SetCell[r=2]{c,m} {\textbf{B}\\ \textit{Repetition}}
    & B1 & Lengthening
    & Repeated characters convey prosodic emphasis; the canonical form collapses the repetition.
    & {\textit{soooo}\,$\to$\,\textit{so}} \\
    & B2 & Iteration
    & Repeated tokens or words; the canonical form retains the repetition for phonemization.
    & {\textit{wait wait}} \\
    \midrule
    \SetCell[r=4]{c,m} {\textbf{C}\\ \textit{Pass-through}}
    & C1 & Eye Direct
    & A phonetic respelling whose surface form already encodes the intended pronunciation.
    & {\textit{luv}\,$\to$\,\textit{love}\\ \vn{\textit{zô}}\,$\to$\,\vn{\textit{vô}}} \\
    & C2 & Regular
    & Standard productive variation of a canonical word.
    & {\textit{droppin}\,$\to$\,\textit{dropping}} \\
    & C3 & Slang/Teen
    & Lexicalized non-standard form whose pronunciation is well established.
    & {\textit{tho}\,$\to$\,\textit{though}\\ \vn{\textit{hong}}\,$\to$\,\vn{\textit{không}}} \\
    & C4 & {Phrasal\\ (Pronounceable)}
    & Multi-letter abbreviation pronounced as a word in its own right.
    & {\textit{rofl}\\ \vn{\textit{GATO}}} \\
    \bottomrule[2pt]
    \end{tblr}
    \caption{Operational definitions and representative examples of the fine-grained categories for non-canonical words.}
    \label{tab:category_definitions}
\end{table*}

\paragraph{Category definitions.}
We use the nine fine-grained categories listed in \Cref{tab:taxonomy}, grouped into the three macro-types, \emph{Reconstruct} (A1--A3), \emph{Repetition} (B1--B2), and \emph{Pass-through} (C1--C4).
The macro-grouping reflects the type of inference required to phonemize a token, rather than its surface transformation.
\Cref{tab:category_definitions} presents the operational definitions and representative examples used during annotation.

\paragraph{Category annotation pipeline.}
Category labels are assigned manually by native-speaker annotators following the operational definitions in \Cref{tab:category_definitions}.
For each $\Gnc \to \Gc$ pair, an annotator inspects the surrounding sentence and selects the single best-fitting category among A1--C4.
Each pair is then independently verified by a second native-speaker annotator, with disagreements resolved through discussion.
All annotators have prior training in NLP or linguistics and are instructed to flag examples where (i) the canonical form is genuinely ambiguous given the context, or (ii) the category assignment is unclear; cases of type (i) are resolved using sentence context as described in \Cref{ssec:task_defs}, and cases of type (ii) default to the more specific category (e.g., A3 rather than A1 when both apply).

\subsection{Phoneme Annotation}
\label{sec:appendix-phoneme-annotation}

Target phonemes $P_c$ are derived from the canonical form $G_c$ using IPA-Dict~\citep{ipadict}.
For English heteronyms (e.g., \textit{lead} as /\textipa{li:d}/ or /\textipa{lEd}/), the correct pronunciation depends on sentence-level context and IPA-Dict alone is insufficient.
Therefore, an annotator with linguistic expertise manually selects the appropriate phoneme sequence based on the surrounding sentence.

\paragraph{Out-of-vocabulary handling.}
Canonical forms not covered by IPA-Dict are manually transcribed following the same IPA conventions.
We measure OOV over unique canonical-form types across all splits, using the corresponding IPA-Dict lexicons
(\texttt{en\_US}, \texttt{vi\_N}, and \texttt{ko});
non-phonemizable and purely symbolic forms are excluded.
Across all canonical types, the OOV rates are 39.4\% (EN), 23.1\% (VI), and 77.6\% (KO). Restricted to non-canonical UGT targets, they are 4.3\% (48/1,110), 7.4\% (175/2,378), and
92.8\% (1,991/2,146), respectively. The high Korean rate largely reflects the mismatch between
IPA-Dict entries and the inflected canonical forms preserved in \benchmark.

\begin{table*}[t]
  \centering
  \small
  \begin{tabular}{lrrcccccccccc}
  \toprule[2pt]
  \multirow{2}{*}[-2pt]{\textbf{Language}} & \multirow{2}{*}[-2pt]{\textbf{Tokens}} & \multirow{2}{*}[-2pt]{\textbf{Cohen's $\kappa$}} & \multicolumn{10}{c}{\textbf{Raw Agreement by UGT Type (\%)}} \\
  \cmidrule(lr){4-13}
  & & & \textbf{Overall} & \textbf{A1} & \textbf{A2} & \textbf{A3} & \textbf{B1} & \textbf{B2} & \textbf{C1} & \textbf{C2} & \textbf{C3} & \textbf{C4} \\
  \midrule
  English    & 1{,}327 & 0.860 & 86.3 & 98.2 & 87.8 & 51.3 & 100.0 & 86.4 & 92.1 & 88.1 & 88.6 & 67.7 \\
  Vietnamese & 2{,}152 & 0.869 & 87.3 & 84.1 & 92.7 & 83.6 & 97.7  & 48.6 & 89.4 & 92.3 & 91.6 & 0.0  \\
  Korean     & 722     & 0.754 & 75.6 & 87.9 & ---  & ---  & 78.8  & 68.1 & 86.4 & 90.9 & 70.9 & 62.5 \\
  \bottomrule[2pt]
  \end{tabular}
  \caption{
  Inter-annotator agreement for phoneme annotations across languages and UGT types.
  While English and Vietnamese show near-perfect agreement overall, Korean exhibits lower agreement due to greater variation in IPA transcription conventions.
  }
  \label{tab:iaa}
  \end{table*}

\paragraph{Inter-annotator agreement.}
To assess phoneme annotation reliability, two annotators independently perform the procedure above, and we compute agreement between the resulting transcriptions.
As shown in \Cref{tab:iaa}, Cohen's $\kappa$ over IPA phoneme sequences reaches 0.86 for English and 0.87 for Vietnamese, with Korean at 0.75.
The Korean score, while lower, remains within the substantial band \citep{landis1977measurement}.
This mainly reflects divergent IPA transcription conventions, particularly differing choices of consonant allophone symbols.

We report the detailed agreement rates by UGT type in \Cref{tab:iaa}.
For English, agreement remains consistently high across most categories, although relatively lower agreement was observed for A3 and C4.
Vietnamese shows a similar overall pattern, with lower agreement for B2 and especially C4.
Notably, the zero agreement for Vietnamese C4 primarily reflects genuine ambiguity in letter-by-letter IPA renderings of acronyms rather than annotation errors.
Overall, these results indicate that the annotation process produces reliable phoneme transcriptions across languages while also revealing categories where pronunciation conventions are inherently more variable.
For the released dataset, we resolve all inter-annotator conflicts through discussion between the two annotators to produce a single adjudicated version.

\section{Content Moderation Analysis}
\label{sec:appendix-moderation}

\begin{table}[t]
    \centering
    \resizebox{\columnwidth}{!}{%
    \begin{tabular}{l|r|rr}
        \toprule[2pt]
        \textbf{Language}
        & \textbf{Sentences}
        & \textbf{Flagged (raw)}
        & \textbf{Flagged (norm)} \\
        \midrule
        English    &  5{,}029 &  568 (11.29\%) &  606 (12.05\%) \\
        Vietnamese & 10{,}847 &  401 (3.70\%)  &  473 (4.36\%)  \\
        Korean     &  2{,}497 &  321 (12.86\%) &  322 (12.90\%) \\
        \midrule
        Total & 18,373 & 1,290 (7.02\%) & 1,401 (7.63\%) \\
        \bottomrule[2pt]
    \end{tabular}%
    }
    \caption{Sentence-level moderation flagged rates on \benchmark, evaluated with OpenAI \texttt{omni-moderation-2024-09-26} on both raw UGT forms and canonical-form normalizations. A sentence is flagged if any of the 13 moderation categories triggers.}
    \label{tab:moderation-rates}
\end{table}

User-generated text often contains harmful or offensive content, and \benchmark inherits this property from its source corpora.
To characterize the toxicity profile of \benchmark and to measure how canonical-form normalization affects moderation signal, we apply OpenAI's \texttt{omni-moderation-2024-09-26} model to every sentence in the corpus, scoring both the raw UGT form and its canonical-form normalization across $13$ moderation categories. 

\subsection{Flagged Rates}
\Cref{tab:moderation-rates} reports sentence-level flagged rates.
Korean shows the highest flagged rate ($12.9\%$), comparable to English ($11.3\%$) and roughly three times higher than Vietnamese ($3.7\%$).
The dominant category is \textit{harassment} in English and Korean, and \textit{violence} in Vietnamese, reflecting source-corpus content rather than language-intrinsic properties.

\subsection{Effect of Normalization}

In general, canonical-form normalization marginally increases the
overall flagged rate.
The binary flag state changes for 385 of 18,373 sentences (2.10\%),
while the aggregate flagged rate increases by only 0.60 percentage
points (7.02\% $\rightarrow$ 7.63\%).
This indicates that most sentences retain the same moderation outcome
under normalization.
The increase is most pronounced in Vietnamese
(3.70\% $\rightarrow$ 4.36\%), while English
(11.29\% $\rightarrow$ 12.05\%) and Korean
(12.86\% $\rightarrow$ 12.90\%) show smaller changes.

\section{LLM-Based Datastore Expansion}
\label{sec:appendix-db-expansion}

This appendix details the LLM-based expansion of the retrieval datastore $\mathcal{D}$ summarized in \Cref{ssec:method-retrieval}.

\subsection{Prompt Template}
\label{sec:appendix-db-prompt}

We prompt GPT-5.4-mini~\cite{singh2025openai} with seed examples from the training set to generate additional plausible $(G_{\text{nc}}, G_c)$ pairs.
The prompt is structured as follows: (i) a short description of UGT and the desired output format, (ii) $k$ seed pairs sampled from the training datastore, and (iii) an instruction to generate $n$ additional pairs that are similar in style but not identical to the seeds.
We use $k$ seed pairs sampled from all training entries of the given UGT type (ranging from a few dozen to over 200 per type depending on training-set density) and set $n \in \{150, 100, 200\}$ for types A1, A2, and A3, respectively. The remaining categories (B1, B2, C1--C4) are populated from the training-set entries only, without LLM-generated augmentation.

The full prompt template is given below: \texttt{\{nc\_type\}}, \texttt{\{type\_desc\}}, \texttt{\{lang\_name\}}, \texttt{\{seed\_lines\}}, and \texttt{\{target\}} are filled per type. The system message instructs the model to act as an expert in the target language's internet slang.

\begin{quote}\small\ttfamily
Here are real examples of \{nc\_type\} type UGT tokens from \{lang\_name\} social media data.

\{nc\_type\} definition: \{type\_desc\}

Existing examples (do NOT repeat these):\\
\{seed\_lines\}

Generate \{target\} MORE \{nc\_type\} tokens commonly used in \{lang\_name\} Twitter/Instagram/Discord/Reddit/texting.\\
Requirements:\\
- Follow the same pattern as \{nc\_type\} definition above.\\
- Each expansion must be the standard \{lang\_name\} form.\\
- Do NOT include any of the examples above.\\
- Do NOT include standard dictionary words as the raw form.\\
- Do NOT include readable acronyms that are pronounced letter by letter.\\
- Cover a wide variety: different topics, registers, formality levels.\\
- Return ONLY a JSON object: \{"raw\_token": "expanded form", ...\}
\end{quote}

\subsection{Filtering Criteria}
\label{sec:appendix-db-filter}

Each generated $(G_{\text{nc}}, G_c)$ candidate is filtered before insertion into $\mathcal{D}$.
We discard candidates that satisfy any of the following conditions:
\begin{itemize}[leftmargin=*]
\item \textbf{Trivial identity}: $G_c = G_{\text{nc}}$, meaning no normalization is needed and inserting the pair would provide no useful retrieval signal.
\item \textbf{Standard token conflict}: $G_{\text{nc}}$ matches a canonical token already present in the training or development set, which would risk over-correcting canonical inputs at inference.
\item \textbf{Test set leakage}: The test set is never inspected during generation or filtering. Candidates whose $G_{\text{nc}}$ happens to coincide with a test non-canonical token are retained only if the same surface form already appears in the training set with a consistent $G_c$.
\end{itemize}

\subsection{Coverage Statistics}

The final datastore $D$ used in our main results combines
training-set entries with LLM-augmented pairs.
The training-only datastore contains
$1{,}413 / 3{,}674 / 1{,}607$ unique non-canonical surface-form
keys for English, Vietnamese, and Korean, respectively.
The LLM-based expansion adds
$491 / 220 / 246$ entries, respectively.

We define test-set coverage as the proportion of non-canonical
target-token occurrences whose surface form $G_{\mathrm{nc}}$
has an exact-match key in the datastore.
Under this definition, the training-only datastore covers
$70.9\%$ ($941/1{,}327$),
$82.9\%$ ($1{,}785/2{,}152$), and
$25.3\%$ ($183/722$)
of the English, Vietnamese, and Korean test targets, respectively.

\section{Training and Implementation Details}
\label{sec:appendix-training}

\paragraph{Backbones.}
We test our approach with two transformer backbones: ByT5-small~\citep{xue2022byt5} (300M parameters, byte-level encoder-decoder, default) and Qwen2.5-0.5B~\citep{qwen2025qwen25technicalreport} (500M parameters, decoder-only causal LM).

\paragraph{Optimization.}
All backbones are trained with AdamW (learning rate $1\!\times\!10^{-4}$, batch size $32$, $10$ epochs) on 4 NVIDIA V100 GPUs.
We select the checkpoint with the lowest development set $\PERnc$.

\paragraph{Compute budget.}
Each of our method's training runs takes approximately $6$~hours on $4$ NVIDIA V100 GPUs. Including baseline retraining, backbone ablation, and frontier-LLM inference, the total compute used for all results reported in the paper is approximately $400$ V100-GPU hours. Frontier-LLM API calls are not included in this figure.

\paragraph{Decoding.}
During inference, the model decodes greedily.
For non-canonical tokens, the model first emits $G_c$ following the \texttt{<Target>} marker, then the phoneme sequence $P_c$ following \texttt{<Phoneme>}; for canonical tokens, $P_c$ is emitted directly without the \texttt{<Target>} step.

\paragraph{Baselines.}
The rule-based baselines are Epitran~\citep{mortensen2018epitran} and viG2P~\citep{vig2p}, both used out-of-the-box with default settings. For viG2P, we use the Viphoneme implementation.\footnote{\url{https://github.com/v-nhandt21/Viphoneme}}
The supervised neural baselines comprise Transformer~\citep{yolchuyeva2020transformer}, ByT5 and ByT5-tiny~\citep{xue2022byt5}, and GBERT~\citep{dong2022neural}; each is trained on the IPA-Dict training split with the same hyperparameters as in its original paper.
The frontier-LLM baselines are GPT-4o, GPT-5.4-mini~\citep{singh2025openai}, Gemini-2.5-flash, Gemini-3-flash, and Claude-Sonnet-4.6, queried via their official APIs (model identifiers: \texttt{gpt-4o}, \texttt{gpt-5.4-mini}, \texttt{gemini-2.5-flash}, \texttt{gemini-3-flash}, and \texttt{claude-sonnet-4-6}; all accessed May~2026).
For the $K=9$ setting, we use one in-context example per available UGT category, yielding nine examples for English and Vietnamese and seven for Korean, where A2 and A3 are absent.
For the $K\in\{0,9,27\}$ shot-scaling study, this corresponds to 37,440 frontier-LLM requests (2,496 test sentences $\times$ 5 models $\times$ 3 shot settings).
The task-relevant baselines are reproduced or adapted following their published descriptions and evaluated with the same \benchmark splits and PER pipeline: ICKR~\citep{han2024improving} uses GPT-4o with in-context knowledge retrieval; R-G2P~\citep{zhao2022r} uses its controlled-noise robustness training; MoNoise~\citep{van2019monoise} (English only) and PolyNorm~\citep{wong2025polynorm} perform lexical normalization before the same word-level ByT5 G2P backend.

\begin{table*}[t]
\centering
\small
\setlength{\tabcolsep}{4pt}
\begin{tabular}{lrrrrrr}
\toprule
& \multicolumn{2}{c}{\textbf{English}} & \multicolumn{2}{c}{\textbf{Vietnamese}} & \multicolumn{2}{c}{\textbf{Korean}} \\
\cmidrule(lr){2-3}\cmidrule(lr){4-5}\cmidrule(lr){6-7}
\textbf{Model} & $\PERc$ & $\PERnc$ & $\PERc$ & $\PERnc$ & $\PERc$ & $\PERnc$ \\
\midrule
ByT5 (300M) & 5.2 & 29.8 & 7.7 & 43.2 & 4.3 & 47.5 \\
Qwen2.5-0.5B (naive FT) & 11.2 & 14.8 & 4.4 & 23.1 & 14.2 & 77.3 \\
ICKR (GPT-4o, reproduced) & 7.7 & 14.9 & 7.9 & 30.8 & 8.2 & 47.9 \\
R-G2P (reproduced) & 18.2 & 53.2 & 6.9 & 59.3 & 4.1 & 42.3 \\
PolyNorm (GPT-4) + ByT5 & 25.8 & 44.5 & 15.6 & 49.3 & 15.5 & \textbf{28.5} \\
MoNoise + ByT5 & 14.2 & 18.0 & -- & -- & -- & -- \\
\midrule
Ours (Qwen2.5-0.5B) & 10.2 & \textbf{14.3} & 4.0 & \textbf{20.0} & 6.0 & 32.4 \\
Ours (ByT5, 300M) & \textbf{4.7} & 22.5 & \textbf{5.9} & 26.7 & 5.1 & 32.0 \\
\bottomrule
\end{tabular}
\caption{Task-relevant and matched-backbone comparisons on \benchmark (PER, \%). Bold marks the best value among the listed systems for each metric. Normalization-first systems can improve $\PERnc$ while substantially increasing $\PERc$.}
\label{tab:appendix-task-baselines}
\end{table*}

\paragraph{Multi-seed stability.}
We rerun both proposed variants with three random seeds and report mean $\pm$ sample standard deviation in \Cref{tab:appendix-stability}. English and Vietnamese are stable; Korean varies more, consistent with its smaller test split and larger unseen-form rate.

\begin{table}[t]
\centering
\small
\setlength{\tabcolsep}{4pt}
\begin{tabular}{llcc}
\toprule
\textbf{Backbone} & \textbf{Lang.} & $\PERc$ & $\PERnc$ \\
\midrule
Qwen2.5-0.5B & EN & $10.07\!\pm\!0.15$ & $14.03\!\pm\!0.31$ \\
& VI & $3.93\!\pm\!0.06$ & $20.03\!\pm\!0.06$ \\
& KO & $4.47\!\pm\!1.39$ & $29.57\!\pm\!2.60$ \\
ByT5 & EN & $4.70\!\pm\!0.10$ & $20.60\!\pm\!2.07$ \\
& VI & $5.83\!\pm\!0.06$ & $27.07\!\pm\!0.32$ \\
& KO & $4.73\!\pm\!0.32$ & $30.17\!\pm\!1.59$ \\
\bottomrule
\end{tabular}
\caption{Three-seed mean $\pm$ sample SD PER (\%).}
\label{tab:appendix-stability}
\end{table}

\paragraph{Paired statistical tests.}
We additionally run two-sided paired bootstrap tests with 10,000 sentence-level resamples on the checkpoints reported in the main tables. \Cref{tab:appendix-bootstrap} reports $\Delta=\text{Ours}-\text{matched baseline}$ for $\PERnc$. All intervals exclude zero, although the English Qwen effect is small.

\begin{table}[t]
\centering
\small
\setlength{\tabcolsep}{4pt}
\begin{tabular}{llrrl}
\toprule
\textbf{Lang.} & \textbf{Backbone} & \textbf{Ours} & \textbf{Base} & \textbf{95\% CI of $\Delta$} \\
\midrule
EN & Qwen & 14.3 & 14.8 & $[-0.8,-0.1]$ \\
EN & ByT5 & 22.5 & 29.8 & $[-8.9,-3.4]$ \\
VI & Qwen & 20.0 & 23.1 & $[-5.4,-1.3]$ \\
VI & ByT5 & 26.7 & 43.2 & $[-18.3,-13.8]$ \\
KO & Qwen & 32.4 & 77.3 & $[-47.9,-43.1]$ \\
KO & ByT5 & 32.0 & 47.5 & $[-18.4,-10.3]$ \\
\bottomrule
\end{tabular}
\caption{Paired-bootstrap $\PERnc$ comparisons (10,000 sentence-level resamples).}
\label{tab:appendix-bootstrap}
\end{table}

\begin{table*}[t]
  \centering
  \small
  \setlength{\tabcolsep}{4pt}
  \begin{tabular}{cll}
    \toprule
    \textbf{Cat.} & \textbf{User (token list)} & \textbf{Assistant (IPA list)} \\
    \midrule
    A1 & \emph{Some / guys / hv / a / bad / game}      & \textipa{"s@m / "gaIz / "h\ae v / @ / "b\ae d / "geIm} \\
    A2 & \emph{bruh / y / u / not / playin}            & \textipa{b\*r2 / "waI / "ju / "nAt / "pleII\ng} \\
    A3 & \emph{naah / im / just / idk / sorry}         & \textipa{"nA / "aIm / "dZ@st / "aI doUnt noU / "sA\*ri} \\
    B1 & \emph{yeah / like / a / loooot}               & \textipa{"j\ae / "laIk / "eI / "lOt} \\
    B2 & \emph{frfr / goodmorning}                     & \textipa{f\*r\textrhookrevepsilon{} f\*r\textrhookrevepsilon{} / "gUd""mO\*rnI\ng} \\
    C1 & \emph{wat / u / kno / bout / hockey}          & \textipa{"w@t / "ju / "noU / @"baUt / "hAki} \\
    C2 & \emph{Wats / goin / on}                       & \textipa{"hw@ts / "goUI\ng / "An} \\
    C3 & \emph{goin / wit / da / crew}                 & \textipa{"goUI\ng / "wIT / D@ / "k\*ru} \\
    C4 & \emph{i / thought / of / u / omg}             & \textipa{"aI / "TOt / @v / "ju / "oU "maI "gAd} \\
    \bottomrule
  \end{tabular}
  \caption{English few-shot example pool: one representative (\emph{user}, \emph{assistant}) pair per UGT category, drawn first when $K=9$. ``/'' separates the per-token entries that the prompt presents on numbered lines.}
  \label{tab:appendix-fewshot-en}
\end{table*}

\section{Frontier LLM Prompt Template}
\label{sec:appendix-prompts}

This section describes the in-context learning setup used to query the frontier-LLM baselines (\Cref{sec:appendix-training}). We use the same chat-message structure across GPT-4o, GPT-5.4-mini, Gemini-2.5-flash, Gemini-3-flash, and Claude-Sonnet-4.6, with decoding parameters \texttt{temperature=0} and \texttt{max\_completion\_tokens=512}.

\paragraph{Message structure.}
For each input sentence $S$, we issue a single chat request consisting of (i) a language-specific system message, (ii) a set of few-shot exchanges, each as a (\emph{user}, \emph{assistant}) pair, and (iii) the query sentence as the final \emph{user} turn:
\begin{quote}\small\ttfamily
[system]\ \ language-specific G2P instructions\\[2pt]
[user]\ \ \ \ 1. tok$_1$ \textbackslash n 2. tok$_2$ \textbackslash n \ldots\\
{[assistant]}\ \ \ 1. ipa$_1$ \textbackslash n 2. ipa$_2$ \textbackslash n \ldots\\
\ldots\ (few-shot pairs)\\[2pt]
[user]\ \ \ \ 1. $w_1$ \textbackslash n 2. $w_2$ \textbackslash n \ldots
\end{quote}
Both input and output use a numbered one-token-per-line format, and responses are parsed with the regex \texttt{\textasciicircum(\textbackslash d+)[.):\textbackslash s]\textbackslash s*(.*)\$}. Each token in a sentence is presented to the model as one numbered line, and the model returns the IPA transcription on the same line number.

\paragraph{Shot scaling.}
The nominal $K \in \{0, 9, 27\}$ settings in \Cref{fig:shot_scaling} correspond to: \emph{$K{=}0$}, with no few-shot pairs; \emph{$K{=}9$}, with one pair per available UGT category; and \emph{$K{=}27$}, with three pairs per available UGT category. Thus, the actual numbers of in-context examples are $0/9/27$ for English and Vietnamese and $0/7/21$ for Korean,
since A2 and A3 do not appear in Korean. Few-shot example pairs are sampled from the training split, drawn once per UGT category, and reused across all queries.

\paragraph{System messages.}
The system message specifies the G2P task and target IPA conventions for each language. We reproduce the English system message below; the Vietnamese and Korean variants follow the same template, written in the target language and referencing language-specific phonological conventions (Vietnamese tone marks, Korean tense consonants and consonant assimilation).

\begin{quote}\small\ttfamily
You are a grapheme-to-phoneme (G2P) expert for English. You will receive a numbered list of tokens from a sentence (possibly containing informal, abbreviated, or slang forms). Output the IPA transcription for each token in the same numbered order.\\
Rules:\\
- Each line of input is ONE token, identified by its number (e.g. `3. idk').\\
- Output exactly one numbered IPA line per token: `3. \textipa{"aI doUnt noU}'.\\
- A token may contain spaces (e.g. a NeMo-expanded handle) --- treat it as one unit.\\
- For abbreviated/informal tokens (e.g. \emph{pls}$\to$\emph{please}, \emph{idk}$\to$\emph{i don't know}, \emph{sooo}$\to$\emph{so}), first determine the standard form, then transcribe.\\
- For repeated chars/words (e.g. \emph{frfr}, \emph{sooooo}), transcribe the normalized form.\\
- Use standard IPA notation.\\
- Output ONLY numbered IPA lines, no explanations or extra text.
\end{quote}

\paragraph{Few-shot example pools.}
For each language and UGT category, we maintain a fixed pool of three (\emph{token list}, \emph{IPA list}) pairs and draw $n$ pairs per available category in order, where $n=1$ for the nominal $K{=}9$ setting and $n=3$ for the nominal $K{=}27$ setting. Because Korean lacks A2 and A3, these settings contain 7 and 21 actual in-context examples for Korean, respectively.
Each pair is presented at the full sentence level so the model also observes that canonical words inside a sentence should be phonemized normally. We illustrate one (\emph{user}, \emph{assistant}) pair per category for English in \Cref{tab:appendix-fewshot-en}.

\section{Downstream TTS-ASR Evaluation}
\label{sec:appendix-tts}

To examine whether token-level PER gains transfer to synthesized speech, we conduct TTS-ASR evaluations in English and Vietnamese using eSpeak NG and Whisper large-v3.

\begin{table}[t]
  \centering
  \small
  \begin{tabular}{lcc}
    \toprule
    \textbf{System} & \textbf{PER}$_{\text{overall}}$\,\dn & \textbf{WER}\,\dn \\
    \midrule
    Canonical (gold IPA) & --- & $0.410$ \\
    GPT-4o ($9$-shot)    & $0.111$ & $0.382$ \\
    Word ByT5            & $0.078$ & $0.368$ \\
    Ours (Qwen 2.5-0.5B) & $0.140$ & $0.436$ \\
    \rowcolor{Blue}
    Ours (ByT5)        & $\mathbf{0.066}$ & $\mathbf{0.359}$ \\
    \bottomrule
  \end{tabular}
  \caption{Token-level overall PER and sentence-level WER on the $100$-sentence English TTS-ASR evaluation. Lower is better. Bold: best per column.}
  \label{tab:appendix-tts-overall}
\end{table}

\begin{table}[t]
  \centering
  \small
  \setlength{\tabcolsep}{4pt}
  \resizebox{\columnwidth}{!}{%
  \begin{tabular}{lccccc}
    \toprule
    \textbf{Type} & \textbf{Can.} & \textbf{Word} & \textbf{GPT-4o} & \textbf{Ours} & \textbf{Ours} \\
                  &               & \textbf{ByT5} & ($9$-shot)     & \textbf{Qwen} & \textbf{ByT5} \\
    \midrule
    A1 & $0.541$ & $0.438$ & $0.481$ & $0.492$ & $\mathbf{0.423}$ \\
    A2 & $0.396$ & $0.466$ & $0.425$ & $0.547$ & $\mathbf{0.379}$ \\
    A3 & $\mathbf{0.431}$ & $0.485$ & $0.462$ & $0.559$ & $0.459$ \\
    B1 & $0.363$ & $\mathbf{0.271}$ & $0.327$ & $0.341$ & $0.303$ \\
    B2 & $0.369$ & $0.301$ & $\mathbf{0.266}$ & $0.372$ & $0.369$ \\
    C1 & $0.392$ & $0.409$ & $0.467$ & $0.469$ & $\mathbf{0.378}$ \\
    C2 & $0.352$ & $0.324$ & $0.356$ & $0.385$ & $\mathbf{0.318}$ \\
    C3 & $0.387$ & $0.309$ & $0.331$ & $0.374$ & $\mathbf{0.313}$ \\
    C4 & $0.456$ & $0.371$ & $\mathbf{0.361}$ & $0.448$ & $0.429$ \\
    \bottomrule
  \end{tabular}}
  \caption{Sentence-level WER by UGT category on the $100$-sentence English TTS-ASR evaluation. Bold: best per row. Ours (ByT5) wins on $5/9$ categories, with the largest gains on Reconstruct (A1--A2) and Pass-through phonetic respelling (C1--C3), where retrieved canonical-form evidence directly resolves the surface.}
  \label{tab:appendix-tts-percat}
\end{table}

\paragraph{Vietnamese evaluation.}
We additionally evaluate $297$ Vietnamese examples. \Cref{tab:appendix-tts-vi} reports sentence-level WER with 95\% bootstrap confidence intervals from 10,000 resamples. Our Qwen2.5-0.5B variant attains the lowest observed WER among predicted-input systems ($0.640$), followed by our ByT5 variant ($0.660$).

\begin{table}[t]
  \centering
  \small
  \setlength{\tabcolsep}{3.5pt}
  \resizebox{\columnwidth}{!}{%
  \begin{tabular}{lccc}
    \toprule
    \textbf{System} & \textbf{WER}\,\dn & \textbf{95\% CI} & \textbf{UGT PER}\,\dn \\
    \midrule
    Gold canonical input & $0.634$ & $[0.607,0.661]$ & $0.000$ \\
    Word ByT5 & $0.667$ & $[0.640,0.694]$ & $0.432$ \\
    GPT-4o (few-shot) & $0.712$ & $[0.687,0.737]$ & $0.456$ \\
    Ours (Qwen 2.5-0.5B) & $\mathbf{0.640}$ & $[0.613,0.666]$ & $\mathbf{0.200}$ \\
    Ours (ByT5) & $0.660$ & $[0.633,0.686]$ & $0.272$ \\
    \bottomrule
  \end{tabular}}
  \caption{Vietnamese downstream TTS-ASR evaluation on $297$ examples. WER confidence intervals use 10,000 bootstrap resamples. Lower is better; bold marks the best predicted-input system.}
  \label{tab:appendix-tts-vi}
\end{table}

\paragraph{Setup.}
We sample $100$ sentences from \benchmark's English test split, stratified by dominant UGT macro-type ($30$ Reconstruct, $20$ Repetition, $50$ Pass-through; mean length $11.7$ tokens, $1{,}169$ valid tokens total) under seed $42$. For each system, the predicted IPA is synthesized with eSpeak NG~1.50 (\texttt{en-us} voice, direct IPA input) and transcribed by OpenAI Whisper large-v3~\citep{radford2023robust} (temperature $0$, fp16). We compute WER between the transcript and the NeMo-normalised gold sentence. Five systems are compared: \textit{Canonical} (gold IPA from $\Gc$), \textit{Word ByT5} (strongest non-retrieval baseline), \textit{GPT-4o (9-shot)}, and our approach with the Qwen2.5-0.5B and ByT5-small backbones.

\paragraph{Caveats.}
For the English evaluation, $N=100$ makes absolute WER differences below roughly $\pm 0.02$ unreliable without a bootstrap interval, which we do not compute. \textit{Canonical} is not a strict upper bound: gold IPA contains length and secondary-stress marks (\textipa{:}, \textipa{"}) that eSpeak NG does not always realize, so the gap to Canonical over-estimates the achievable improvement.

\paragraph{Overall WER.}
\Cref{tab:appendix-tts-overall} reports overall sentence-level WER. Ours (ByT5) achieves the lowest WER ($0.359$), below Word ByT5 ($0.368$, $-0.009$) and GPT-4o ($0.382$, $-0.023$). The downstream ranking matches the token-level PER ranking across all four trained systems, indicating that PER on \benchmark{} is a faithful proxy for downstream intelligibility on this evaluation. Ours (Qwen) underperforms ($0.436$), consistent with its higher token-level PER ($0.140$).

\paragraph{Per-UGT-type WER.}
\Cref{tab:appendix-tts-percat} breaks WER down by UGT category. Ours (ByT5) wins on $5/9$ categories (A1, A2, C1, C2, C3), with the largest gains on recoverable shortenings: A2 ($-0.087$ vs.\ Word ByT5) and A1 ($-0.015$). Where our approach does not win, the failure mode is interpretable: A3 (phrasal abbreviation, e.g., \textit{idk}, \textit{nvm}) requires multi-word expansion that the single-token retrieval store does not cover; B-type repetition is already handled by the byte-level encoder of Word ByT5; and C4 (phrasal acronyms) benefits from GPT-4o's broader world knowledge. Categories that benefit most from retrieval at the token level (A1--A2, C1--C3) are also where our model wins downstream, corroborating that the taxonomy localizes the same difficulty axis at both levels.

\end{document}